\documentclass[runningheads]{llncs}
\usepackage{graphicx}
\usepackage{amsmath,amssymb} 
\usepackage{color}
\usepackage[width=122mm,left=12mm,paperwidth=146mm,height=193mm,top=12mm,paperheight=217mm]{geometry}
\usepackage{rotating}

\usepackage{booktabs}
\usepackage{xspace}

\usepackage{tabularx}

\newcommand*{\etal}{\textit{et al.}\@\xspace} 
\newcommand*{\eg}{\textit{e.g.}\@\xspace}     
\newcommand*{\ie}{\textit{i.e.}\@\xspace}     

\newcommand{\sota}{state-of-the-art }

\def\ismpl{k}
\def\nsmpl{K}
\def\smpl{s}
\def\smpli{\smpl_\ismpl}
\def\trgt{t}
\def\trgti{\trgt_\ismpl}

\def\wgts{w}

\def\scale{\rho}

\def\RealN{\mathbb{R}}

\def\st{\text{s.t.}}
\DeclareMathOperator*{\argmin}{arg\,min}

\DeclareMathOperator{\proj}{proj}

\DeclareMathOperator{\TGV}{TGV}
\DeclareMathOperator{\ReLU}{ReLU}

\newcommand{\set}[1]{\left\lbrace #1 \right\rbrace}
\newcommand{\norm}[1]{\left\lVert #1\right\rVert}
\newcommand{\sclrprd}[2]{\left\langle #1, #2 \right\rangle}

\begin{document}
\pagestyle{headings}
\mainmatter

\title{ATGV-Net: Accurate Depth Super-Resolution} 

\titlerunning{ATGV-Net: Accurate Depth Super-Resolution}

\authorrunning{Riegler, R\"uther, Bischof}

\author{Gernot Riegler, Matthias R\"uther, Horst Bischof}


\institute{Institute for Computer Graphics and Vision,\\
	Graz University of Technology\\
	\email{ \{riegler, ruether, bischof\}@icg.tugraz.at}
}

\maketitle

\begin{abstract}
In this work we present a novel approach for single depth map super-resolution.
Modern consumer depth sensors, especially Time-of-Flight sensors, produce dense depth measurements, but are affected by noise and have a low lateral resolution.
We propose a method that combines the benefits of recent advances in machine learning based single image super-resolution, \ie deep convolutional networks, with a variational method to recover accurate high-resolution depth maps.
In particular, we integrate a variational method that models the piecewise affine structures apparent in depth data via an anisotropic total generalized variation regularization term on top of a deep network.
We call our method \emph{ATGV-Net} and train it end-to-end by unrolling the optimization procedure of the variational method.
To train deep networks, a large corpus of training data with accurate ground-truth is required.
We demonstrate that it is feasible to train our method solely on synthetic data that we generate in large quantities for this task.
Our evaluations show that we achieve \sota results on three different benchmarks, as well as on a challenging Time-of-Flight dataset, all without utilizing an additional intensity image as guidance.
\end{abstract}

\section{Introduction}
Over the last decade depth sensors have entered the mass market which substantially improved in package size, energy consumption and price.
This made depth data an interesting and important auxiliary input for computer vision tasks, for example in pose estimation~\cite{girshick11,shotton11}, or scene understanding~\cite{gupta14}.
However, current sensors are limited by physical and manufacturing constraints.
Hence, depth outputs are affected by degenerations due to noise, quantization and missing values, and typically have a low resolution.

To alleviate the use of depth data, recent methods focus on increasing the spatial resolution of the acquired depth maps.
A common approach to tackle this problem is to utilize a high-resolution intensity image as guidance~\cite{ferstl13,kwon15,park11}.
These methods are motivated by the statistical co-occurrences of edges in intensity images and discontinuities in depth.
In practical scenarios, however, a depth sensor is not always accompanied by an additional camera and the depth map has to be projected to the guidance image, which is also problematic due to noisy depth measurements.
Therefore, approaches that solely rely on the depth input for super-resolution are becoming popular~\cite{aodha12,ferstl15,hornacek13}.

In contrast to super-resolution methods for depth data, machine learning based methods for natural images~\cite{dong14,schulter15,timofte13,timofte14} are advancing rapidly and achieve impressive results on standard benchmarks.
Those methods learn a mapping from a low-resolution input space to a plausible and visually pleasing high-resolution output space.
The inference is performed for small, overlapping patches of the image independently, and are then averaged for the final output.
This is not optimal for depth data, as it is characterised by textureless, piece-wise affine regions that have sharp depth discontinuities.
In contrast, variational methods are especially suited for this task, because the aforementioned prior information can be exploited in the model's regularization term.
A prominent example is the total generalized variation (TGV)~\cite{bredies10} that is for example utilized in~\cite{ferstl13}.

In this work we propose a method that combines the advantages of data-driven methods and energy minimization models by combining a deep convolutional network with a powerful variational model to compute an accurate high-resolution output from a single low-resolution depth map input.
Deep networks recently demonstrated impressive capabilities in single-image super resolution~\cite{kim15a}.
We utilize a similar architecture for our network, but instead of just producing the refined depth map as output, we design the network to additionally predict the locations of the depth discontinuities in the high-resolution output space.
Both outputs are then used as input for a variational model to refine the high-resolution estimate.
The variational model uses an anisotropic TGV pairwise regularization that is weighted by the network output.
To integrate the variational method into our network and learn the joint model end-to-end, we unroll all computation steps of the primal-dual optimization scheme~\cite{chambolle11} that is used for inference with layers of a deep network.
Therefore, we name our method \emph{ATGV-Net}.
Finally, we deal with the problem of obtaining accurate ground-truth data for training.
The training of deep networks requires a large corpus of data.
We demonstrate that we can train our model entirely on synthetic depth data that we generate in large quantities and obtain \sota results on four different benchmark datasets.

Our contributions can be summarized as follows:
(i) We integrate a variational model with anisotropic TGV regularization into a deep network by unrolling the optimization steps of the primal-dual algorithm~\cite{chambolle11} and train the whole model end-to-end (see Sec.~\ref{sec:method}).
(ii) We demonstrate that our joint model can be trained entirely on synthetic data for single depth map super-resolution (see Sec.~\ref{sec:training_data}).
(iii) Finally, we show that our method improves upon \sota results on four different benchmark datasets (see Sec.~\ref{sec:eval_mb_ls}-\ref{sec:eval_tofmark}).

\section{Related Work}

\noindent\emph{Depth Super-Resolution}
In general, the work on super-resolution is roughly divided in approaches that use a series of aligned images to produce a high-resolution output, and single image super-resolution, \ie approaches that use only one low-resolution image as input.
We focus in this related work on the latter as our method falls into this category.

Natural images often contain repetitive structures and therefore, a patch might be visible on different scales within the same image.
Glasner~\etal~\cite{glasner09} exploit this knowledge in their seminal work.
For each image patch they search similar patches across various scales in the image and combine them for a high-resolution estimate.
A similar idea is employed for depth data by Horn\'{a}\v{c}ek~\etal~\cite{hornacek13}, but instead of reasoning about 2D patches, they reason in terms of patches containing 3D points.
The 3D points of the depth map patch can be translated and rotated with six degree of freedom to find related patches within the same depth map.
Aodha~\etal~\cite{aodha12} search for similar patches not within the same image, but in an ancillary database and they formulate a Markov Random Field~(MRF) that enforces smooth transition between the candidate high-resolution patches.

More recently, machine learning approaches have become popular for single image super-resolution.
They achieve higher accuracy and are at the same time more efficient in testing, because they do not rely on a computational intensive patch search.
Sparse coding approaches~\cite{yang10,zeyde10} learn dictionaries for the low- and high-resolution domains that are coupled via a common encoding.
To increase the inference speed, Timofte~\etal~\cite{timofte13} replace the $\ell_1$ norm in the sparse coding step with the $\ell_2$ norm, which can be solved in closed form and replace a single dictionary by man smaller sub-dictionaries to improve accuracy.
In \cite{schulter15}, Schulter~\etal substitute the flat code-book of sparse coding methods with a random regression forest.
A test patch traverses the trees of the forest and each leaf node stores regression coefficients to predict a high-resolution estimate.
Deep learning based approaches recently showed very good results for single image super-resolution, too.
Dong~\etal~\cite{dong14} train a convolutional network of three layers. 
The input to the network is the bilinear upsampled low-resolution image and the network is trained with the Euclidean loss on the network output and the corresponding ground-truth high-resolution image.
This idea was substantially improved by Kim~\etal~\cite{kim15a}.
They train a deep network with up to 20 convolutional layers with filters of size $3 \times 3$ and therefore, increasing the receptive field to $41 \times 41$ pixel from $15 \times 15$ pixels of the network in \cite{dong14}.
Further, the network does not output directly the high-resolution estimate, but the residual to the pre-processed input image, aiding training of the very deep networks~\cite{he15}.

These learning based methods have mainly been applied to color images, where a huge amount of training data can be easily obtained.
In contrast, large datasets with dense, accurate depth maps have only very recently become available, \eg~\cite{handa16}.
Therefore, most methods for depth map super-resolution are not based on machine learning, but utilize a high-resolution intensity image as guidance.
One of the first works in this direction is by Diebel and Thrun~\cite{diebel05}.
They apply a MRF for the upsampling task and weight their smoothness term according to the gradients of the guidance image.
Yang~\etal~\cite{yang07} propose an approach based on a bilateral filter that is iteratively applied to estimate a high-resolution output map. 
Park~\etal~\cite{park11} present a least-squares method, that incorporates edge aware weighting schemes in the regularization term of their formulation.
A more recent approach of Ferstl~\etal~\cite{ferstl13} utilizes a variational framework for image guided depth upsampling, where they also use the total generalized variation~\cite{bredies10} as regularization term.
One of the few machine learning based approaches for depth map super-resolution is by Kwon~\etal~\cite{kwon15}.
They collect their own training data using KinectFusion~\cite{izadi11} and facilitate sparse coding with an additional multi-scale approach and an advanced edge weighting term, that emphasizes intensity edges corresponding to depth discontinuities.
Ferstl~\etal~\cite{ferstl15} use sparse coding with dictionaries trained on the $31$ synthetic depth maps of \cite{aodha12} to predict the depth discontinuities in the high-resolution domain from the low-resolution depth data.
Those edge estimates are then used in an anisotropic diffusion tensor of their regularization term.

\noindent\emph{Deep Network Integration of Energy Minimization Methods}
Energy minimization methods, such as Markov Random Fields (MRFs), or variational methods have a wide range of applications in computer vision. 
They consist of unary terms, for example the class likelihood of a pixel for semantic segmentation, or the depth value in depth super-resolution, and pairwise terms, which measure the dependencies on neighbouring pixels.
Recently, the integration of those models into deep networks gained a lot of attention, as deep networks jointly trained with energy minimization methods achieve excellent results.
For example, Tompson~\etal~\cite{tompson14a} propose the joint training of a convolutional network and a MRF for human pose estimation.
The MRF is realized by very large convolutional filters to model the pairwise interactions between joints and can be interpreted as one iteration of loopy belief propagation.
In \cite{chen15,schwing15} the authors show how to compute the derivative with respect to the mean field approximation~\cite{kraehenbuehl12} in MRFs.
This allows end-to-end learning and improves results for instance in semantic segmentation.
Similarly, Zheng~\etal~\cite{zheng15} show that the computation steps of the mean field approximation can be modeled by operations of a convolutional network and unroll the iterations on top of their network.

While the latter approaches for semantic segmentation are designed for a discrete label space, the variational approach by Ranftl and Pock~\cite{ranftl14} has a continuous output space.
They show that the gradient of a loss function can be back-propagated through the energy functional of a variational method by implicit differentiation, if the functional is smooth enough.
This approach has been extended for depth denoising and upsampling by Riegler~\etal~\cite{riegler15}.
Recently, Ochs~\etal~\cite{ochs15} propose a technique that allows the back-propagation through non-smooth energy functionals using Bregman proximity functions~\cite{chambolle15}, but did not demonstrate the use in combination with deep networks. 

Our approach utilizes a variational method on top of a deep network, but instead of implicitly differentiating the energy functional as in \cite{ranftl14,riegler15} we unroll every step of an exact optimization scheme~\cite{chambolle11}, in the spirit of~\cite{domke12}.
This has two major advantages: 
First, we can incorporate stronger pairwise regularization terms and second, the optimization gets more robust, allowing the successful training of deeper networks. 
This is similar to \cite{zheng15}, but instead of the mean field approximation, we unroll the steps of the primal-dual algorithm by Chambolle and Pock~\cite{chambolle11}, which converges to the global optimal solution of the convex energy functional.
For parametrizing the variational method we use a $10$ layer deep network of $3 \times 3$ convolutions, and train on the residual similarly to \cite{kim15a}.
Additionally, we train the network to predict the depth discontinuities in the high-resolution output space. 
This output is used to weight the pairwise regularization term of the variational part.
Finally, we demonstrate that we can train a deep network for this task by rendering synthetic depth maps in large quantities with a ray-caster running on the GPU.

\section{ATGV-Net}\label{sec:method}

In this section we describe our method that takes a single low-resolution, probably noisy depth map as input and computes a high-resolution output.
We first introduce the notation used throughout this work and then detail our variational model, how we integrate it on top of a deep network and finally the network itself.

In the remainder of this work we denote the low-resolution depth map input as $\smpli^\text{(lr)} \in \RealN^{M \times N}$.
Further, for training we assume that we have for each input sample an accurate, high-resolution ground-truth depth map $\trgti \in \RealN^{\scale M \times \scale N}$, where $\scale > 1$ is the given upsampling factor.
The only preprocessing step in our method is a bilinear upsampling of the low-resolution input depth map $\smpli^\text{(lr)}$ to the size of the ground-truth target depth map.
We denote this mid-level representation of the input as $\smpli \in \RealN^{\scale M \times \scale N}$.

Given a training set $\set{(\smpli, \trgti)}_{\ismpl=1}^{\nsmpl}$ of $\nsmpl$ training pairs we follow \cite{ranftl14,riegler15} and formulate the training task as the following bi-level optimization problem: 
\begin{align}
  \tag{HL}
  & \min_{w} \frac{1}{\nsmpl} \sum_{\ismpl=1}^\nsmpl L(u^*(f(\wgts, \smpli)), \trgti)
  \label{eq:higher_level} \\
  \tag{LL}
  & \st \quad u^*(f(\wgts, \smpli)) = \argmin_{u} E(u; f(\wgts, \smpli)).
  \label{eq:lower_level}
\end{align}
This optimization problem has an intuitive interpretation: 
In the higher-level problem \eqref{eq:higher_level} we want to minimize some weights $w$, such that the minimizer $u^*$ of the energy functional $E$ in the lower-level problem \eqref{eq:lower_level}, which is parameterized by a learnable function $f$, achieves a low loss $L$ over all training samples.
We provide more details on the energy functional and on the parametrization in Sec.~\ref{sec:pd_as_rnn} and  Sec.~\ref{sec:network_architecture}, respectively.
For the loss we only impose the restriction that we can compute the gradient with respect to $u^*$.
For the remainder of this work we will use the Euclidean loss:
\begin{align}
  L(u^*(f(\wgts, \smpli)), \trgti) = \norm{u^*(f(\wgts, \smpli)) - \trgti}_2^2 .
  \label{eq:loss}
\end{align}
The authors of \cite{ranftl14,riegler15} have proven that the bi-level optimization problem can be solved by implicit differentiation, if certain assumptions for the energy functional $E$ hold.
Namely, $E$ has to be strongly convex, twice differentiable with respect to $u$ and once differentiable with respect to $f$.
Further, the gradient of $f$ has to be computable with respect to $\wgts$.
The last constraint is satisfied by construction since the parametrization $f$ is realized by a deep network.
However, the first constraints drastically limit the choice of energy functionals and therefore, the authors of~\cite{ranftl14,riegler15} had to design smooth approximations.
In the following we show that this constraints can be eliminated by unrolling the optimization steps of the lower-level problem~\eqref{eq:lower_level} on top of a deep network, similar to~\cite{zheng15}.

\begin{figure}[t]
  \center
  \includegraphics[width=\textwidth]{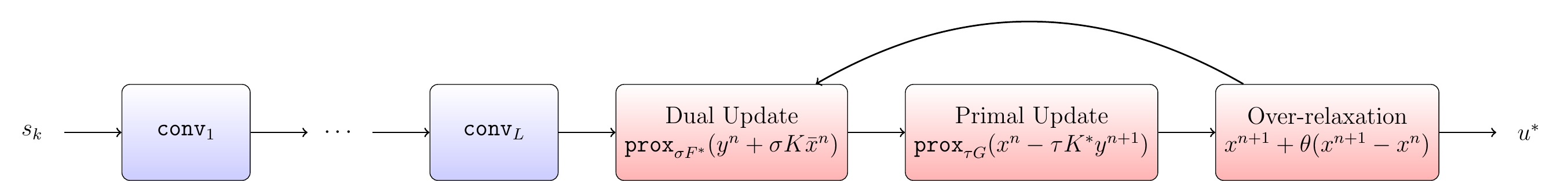}
  \caption{
    Our model consists of a deep convolutional network with $L = 10$ layers (blue rectangles) that predicts a first high-resolution depth map and depth discontinuities.
    The output of the network is then feed to an unrolled primal-dual optimization algorithm (red rectangles) realized by operations in a deep network that further refines the result.
    This enables us to train the joint model end-to-end.
  }
  \label{fig:pd_as_rnn}
\end{figure}

\subsection{Unrolling the Optimization}\label{sec:pd_as_rnn}
For the energy functional we have the requirement that it should refine the initial high-resolution depth estimate.
Therefore, we use a $\TGV_2$-$\ell_2$ variational model~\cite{bredies10} that favors the piecewise affine surfaces apparent in depth maps.
In addition, we incorporate an anisotropic diffusion tensor~\cite{ranftl12,werlberger09} into the regularization and name our model \emph{ATGV-Net}.
The optimization of the energy functional in conjunction with a guidance intensity image already provides good results for depth super-resolution~\cite{ferstl15}.
In the following we demonstrate, how we can significantly improve the model by parametrizing the energy functional by a deep network and learn it end-to-end by unrolling the optimization procedure.

In general, our energy functional consists of a pairwise regularization term $R$ and an $\ell_2$ data term:
\begin{align}
  E(u; f(\wgts, \smpli)) = R(u, h(\wgts_h, \smpli)) + \frac{e^{\wgts_\lambda}}{2} \norm{u - g(\wgts_g, \smpli)}_2^2 \,.
  \label{eq:energy}
\end{align}
The functional is parameterized by a function $f(\wgts, \smpli) = [h(\wgts_h, \smpli), \wgts_\lambda, g(\wgts_g, \smpli)]^T$ that has learnable weights $\wgts$ and takes the mid-resolution depth map $\smpli$ as input.
The functions $h$ and $g$ are realized as a single deep network and described in Sec.~\ref{sec:network_architecture}.
The parameter $\wgts_\lambda$ controls the trade-off between data and regularization term and is also learned. 
We take the exponential of $\wgts_\lambda$ to ensure convexity of the energy functional.
For the pairwise regularization term we utilize the total generalized variation (TGV)~\cite{bredies10} of second order that favors piecewise affine solutions and is therefore ideal for depth maps:
\begin{align}
  R(u, h(\wgts_h, \smpli)) = \min_v \alpha_1 \norm{T(h(\wgts_h, \smpli)) (\nabla_u u - v)}_1 + \alpha_0 \norm{\nabla_v v}_1 \,,
\end{align}
where $\alpha_0$ and $\alpha_1$ are user defined parameters.
In the regularization term, an anisotropic diffusion tensor $T$ enforces a low degree of smoothness across depth discontinuities and vice versa, more smoothness in homogeneous regions.
This anisotropic diffusion tensor is based on the Nagel-Enkelmann operator~\cite{nagel86}:
\begin{align}
  T(h(\wgts_h, \smpli)) = \exp(-\beta \norm{h(\wgts_h, \smpli)}_2^\gamma) n n^T + n_\perp n_\perp^T \,,
\end{align}
with $\beta$ and $\gamma$ being adjustable parameters weighting the magnitude and sharpness of the tensor.
The gradient normal of $h$ is given by
\begin{align}
  n = \frac{h(\wgts_h, \smpli)}{\norm{h(\wgts_h, \smpli)}_2} \,,\quad n_\perp \cdot n = 0 \,. 
\end{align}

To optimize this energy functional we chose the first-order primal-dual algorithm by Chambolle and Pock~\cite{chambolle11}, as it guarantees fast convergence.
To apply the optimization algorithm, we first reformulate Eq.~\eqref{eq:energy} as saddle-point problem with dual variables $p, q$ as
\begin{align}
  &\min_{u, v} \max_{p, q} \alpha_1 \sclrprd{T(h(\wgts_h, \smpli)) (\nabla_u u - v)}{p} + \alpha_0 \sclrprd{\nabla_v v}{q} + \frac{e^{\wgts_\lambda}}{2} \norm{u - g(\wgts_g, \smpli)}_2^2 \\
  &\st\, p \in \set{p \in \RealN^{2 \times \scale M \times \scale N} \mid \norm{p_{:, i, j}}_2 \le 1}, q \in \set{q \in \RealN^{4 \times \scale M \times \scale N} \mid \norm{q_{:, i, j}}_2 \le 1} \,,
\end{align}
where $\nabla_u$ and $\nabla_v$ denote operators in the discrete setting that compute the forward differences of $u$ and $v$.
A single iteration of the optimization procedure to obtain $u^*$ is then given by:
\begin{align}
  p^{n+1} &= \proj(p^n + \sigma_p \alpha_1 (T(h(\wgts_h, \smpli)) (\nabla_u \bar{u}^n - \bar{v}^n))) \label{eq:p_ascent}\\
  q^{n+1} &= \proj(q^n + \sigma_q \alpha_0 \nabla_v \bar{v}^n) \label{eq:q_ascent}\\
  u^{n+1} &= \frac{u^n + \tau_u (\alpha_1 \nabla_u^T T(h(\wgts_h, \smpli)) p^{n+1} + e^{w_\lambda} g(\wgts_g, \smpli))}{1 + \tau_u e^{w_\lambda}} \label{eq:u_descent} \\
  v^{n+1} &= v^n + \tau_v (\alpha_0 \nabla_v^T q^{n+1} + \alpha_1 T(h(\wgts_h, \smpli)) p^{n+1}) \label{eq:v_descent} \\
  \bar{u}^{n+1} &= u^{n+1} + \theta (u^{n+1} - {u}^n) \label{eq:u_or} \\
  \bar{v}^{n+1} &= v^{n+1} + \theta (v^{n+1} - {v}^n) \label{eq:v_or} \,,
\end{align}
with $u^0 = g(\wgts_g, s_k)$, $v^0, p^0, q^0 = 0$, $\sigma_p, \sigma_q, \tau_u, \tau_v > 0$, $\theta \in [0, 1]$, and $\proj(p) = \tfrac{p}{\max(1, \norm{p}_2)}$ is the point-wise projection to the unit hyper-sphere:

The key observations are:
(i) The single computation steps in this optimization algorithm can be realized by operations of a deep network, \ie individual network layers, and (ii) given a fixed number of iterations, the algorithm can be unrolled like a recurrent neural network, similar to~\cite{zheng15}.
This allows us to use the back-propagation algorithm to train the optimization procedure, \ie all hyper-parameters, jointly with the parametrization, \ie the deep network.
See Figure~\ref{fig:pd_as_rnn} for a visualization of the concept.
In the following we detail how the individual computation steps are realised within our model.
We provide a graphical representation of a single iteration of the optimization procedure in terms of deep network operations in the supplemental material.

\emph{Dual Update}
The gradient ascent of the dual variables in Eq.~\eqref{eq:p_ascent} and \eqref{eq:q_ascent} consists of scalar multiplication, point-wise addition and multiplication, the gradient operators $\nabla_u, \nabla_v$, and the projection.
The scalar multiplication and the point-wise operations are trivial operations and are implemented in most deep learning frameworks.
The $\nabla$-operator is basically a convolution with two filters, $\nabla_x = [-1, 1]$ and $\nabla_y = [-1, 1]^T$.
Therefore, it can be implemented with a standard convolutional layer that has fixed filter coefficients.
Additionally, we have to ensure a reflecting padding of the layer input, \ie Neumann boundary conditions.
Finally, the $\proj$-operator is a composition of a point-wise division, a $\max$-operator and the $\ell_2$ norm.
We implemented the $\max$-operator as shifted $\ReLU$, and the $\ell_2$ norm as custom layer.

\emph{Primal Update}
The gradient descent of the primal variables in Eq.~\eqref{eq:u_descent} and \eqref{eq:v_descent} consists of similar operations as the dual update, and therefore, can be implemented with the same building blocks.
Additional operators are $\nabla_u^T, \nabla_v^T$.
These operators are defined as $\nabla^T p = \nabla_x p_x + \nabla_y p_y$.
From this definition we can see that this operation can again be implemented with a convolutional layer that has fixed filter coefficients.
However, we have to ensure a negative symmetric padding of the layer input, \ie Dirichlet boundary conditions.

\emph{Over-Relaxation}
The over-relaxation step of the primal variables in Eq.~\eqref{eq:u_or} and Eq.~\eqref{eq:v_or} can be simplified to a weighted sum of two terms, \ie $\bar{u} = (1 + \theta) u^{n+1} - \theta u^n$ and $\bar{v} = (1 + \theta) v^{n+1} - \theta v^n$.

\subsection{Parametrization}\label{sec:network_architecture}

\begin{figure}[tb]
  \center
  \includegraphics[width=\textwidth]{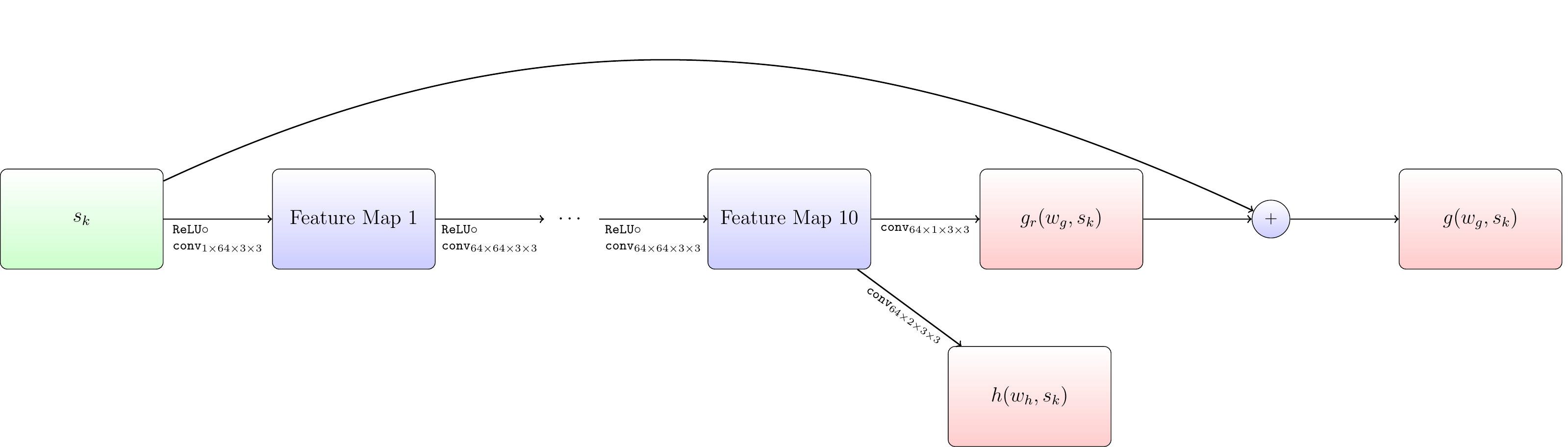}
  \caption{
    Overview of our deep network architecture. 
    Our network consists of $10$ convolutional layers with $3 \times 3$ filters and $64$ feature maps in the hidden layers (blue rectangles).
    The input to the network (green rectangle) is the mid-resolution depth map and the output is (i) the residual that after adding to the mid-resolution input produces the high-resolution estimate $g(\wgts_g, \smpli)$ and (ii) the estimates of the depth discontinuities in the high-resolution output $h(\wgts_h, \smpli)$ (red rectangles).
  }
  \label{fig:network_architecture}
\end{figure}

After we have described the variational model and how to integrate it on top of a deep network, we now detail the parametrization functions $h(\wgts_h, \smpli)$ and $g(\wgts_g, \smpli)$.
Inspired by the recent success in single image super-resolution for color images~\cite{kim15a}, we implement $g(\wgts_g, \smpli)$ as a deep convolutional neural network with $10$ convolutional layers.
Each convolutional filter has the size of $3 \times 3$ and each hidden layer of the network has $64$ feature maps.
As $g(\wgts_g, \smpli)$ is used in the data term of our energy functional it should provide a good initial estimate of the high-resolution depth map.
However, the output of this network is not the estimate of the high-resolution depth map itself, but the residual $g_r(\wgts_g, \smpli)$, such that $g(\wgts_g, \smpli) = g_r(\wgts_g, \smpli) + \smpli$.
Learning the residual instead of the full output aids the training procedure of the network~\cite{kim15a}, and has been applied before in other super-resolution methods~\cite{schulter15,timofte13,timofte14}.

The parameterization function $h(\wgts_h, \smpli)$ is used for weighting the pairwise regularization term.
As we argued before, the regularization should be small near depth discontinuities and high in smooth areas.
Therefore, we implemented $h(\wgts_h, \smpli)$ as an additional network output of size $2 \times \scale M \times \scale N$ and train it to estimate the gradient of the high-resolution target $\nabla \trgti$.
This method has two benefits:
First, we get more accurate estimates for the depth discontinuities than what we would get from the gradient of the high-resolution estimate $g(\wgts_g, \smpli)$.
Secondly, the joint training of both objectives in a single deep network improves the performance of both tasks, because the weights $\wgts_h$ and $\wgts_g$ share the majority of parameters and only the parameters of the last layer, the output, differ.
A graphical depiction of our deep network parametrization is shown in Figure~\ref{fig:network_architecture}.

\subsection{Training}
In the previous sections we presented the description of our model.
In this section we detail how we train it given a large set of training samples $\set{(\smpli, \trgti)}_{\ismpl=1}^\nsmpl$.
The training procedure is two-fold:
In a first step we initialize the deep convolutional network, \ie the functions $g$ and $h$.
Therefore, we train the network by mini-batch gradient descent with momentum term on the following loss function:
\begin{align}
  L_p(\set{(\smpli, \trgti)}_{\ismpl=1}^\nsmpl) = \frac{1}{\nsmpl} \sum_{\ismpl=1}^\nsmpl \norm{g(\wgts_g, \smpli) - \trgti}_2^2 + \norm{h(\wgts_h, \smpli) - \nabla \trgti}_2^2 .
\end{align}
In the following evaluations we set the learning rate to $0.001$ and the momentum parameter to $0.9$ for the initializing of the network.
With this setting we train the network for $30$ epochs on non-overlapping patches of size $32 \times 32$ pixel.

In the second step of the training procedure we add the unrolled primal-dual optimization algorithm as introduced in Sec.~\ref{sec:pd_as_rnn} on top of the network.
Then, we train the joint model end-to-end on the Euclidean loss stated in Eq.~\eqref{eq:loss} with mini-batch gradient descent.
We set the learning rate to $0.001$ and the momentum parameter $0.9$ to train the whole model for $5$ epochs on non-overlapping patches of size $128 \times 128$ pixel.
In contrast to the method of implicit differentiation~\cite{ranftl14,riegler15}, our method is still robust if we use a high learning rate, and as a consequence converges in fewer training iterations.
Further, it enables us to optimize the parameter $\wgts_\lambda$, as well ass all hyper-parameters of the optimization procedure.

\section{Evaluation}\label{sec:evaluation}
In this section we present an exhaustive experimental evaluation of the proposed \emph{ATGV-Net}.
First, we show how we generate a huge amount of training data with accurate ground-truth needed to train the deep network.
Then, we demonstrate evaluation results on four standard benchmark datasets for depth map super-resolution: 
Following~\cite{aodha12,ferstl15,hornacek13}, we evaluate our method on the noise-free Middlebury disparity maps \emph{Teddy}, \emph{Cones}, \emph{Tsukuba} and \emph{Venus}.
Additionally, we show results for the Laserscan dataset as proposed in~\cite{aodha12}.
In a second evaluation we compare our results on the noisy Middlebury 2007 dataset as proposed in~\cite{park11} and finally, we demonstrate the real-world applicability of our method on the challenging ToFMark dataset~\cite{ferstl13}.

We set the initial parameters of our model to $\alpha_1=17$, $\alpha_0=1.2$ for the regularization term, $\beta=9$, $\gamma=0.85$ for the anisotropic diffusion tensor, and $\wgts_\lambda=0.01$ for all experiments.
Further, we fix the number of iterations of the primal-dual algorithm to $10$.

\subsection{Training Data}\label{sec:training_data}

\begin{figure}[tb]
  \center
  \begin{minipage}{0.4\textwidth} \centering
    \includegraphics[width=0.45\textwidth]{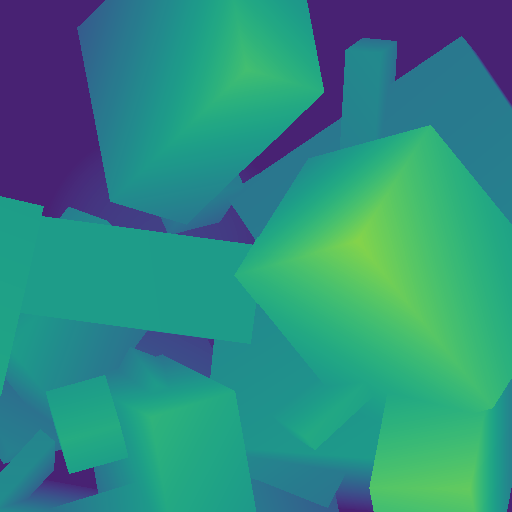}
    \includegraphics[width=0.45\textwidth]{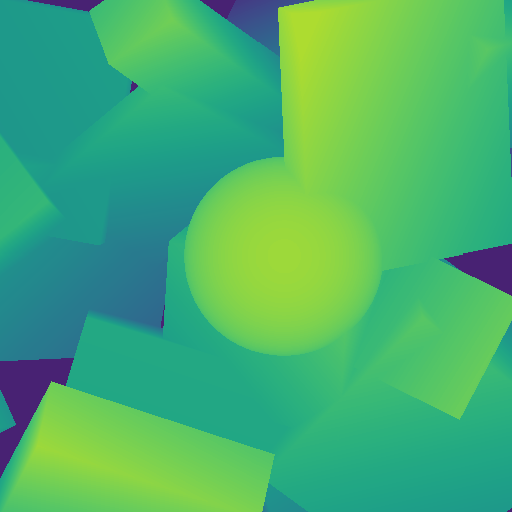}
    {\tiny (a) rendered high-resolution ground-truth}
  \end{minipage}
  \begin{minipage}{0.4\textwidth} \centering
    \includegraphics[width=0.45\textwidth]{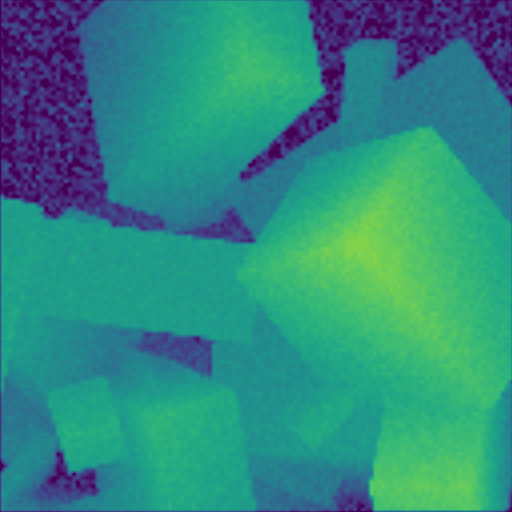}
    \includegraphics[width=0.45\textwidth]{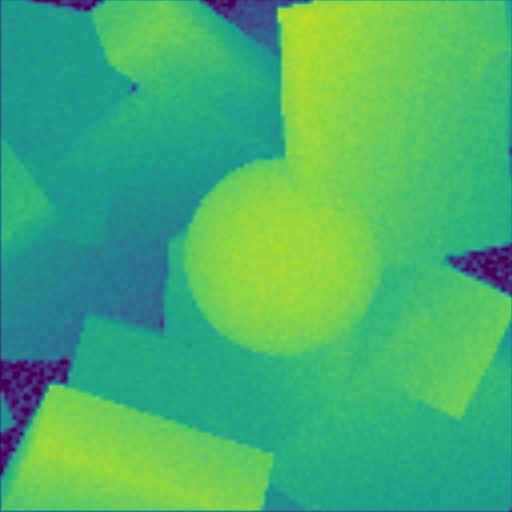}
    {\tiny (b) corresponding mid-resolution input}
  \end{minipage}
  \caption{
    Examples of our generated depth maps.
    (a) visualizes the high-resolution ground-truth data.
    By resampling those depth maps with a scale factor $\scale$ and adding depth dependent noise we create the low-resolution input.
    (b) shows the mid-resolution input, which is the bilinear upsampled low-resolution data.
    Best viewed magnified in the electronic version.
  }
  \label{fig:train_data}
\end{figure}

One challenge in training very deep networks is the need for a huge amount of training data.
In~\cite{aodha12,ferstl15} the authors use a small set, \ie $31$ depth maps, of synthetic rendered images for training and in~\cite{riegler15} the authors trained and tested their method on the synthetic New Tsukuba dataset~\cite{martull12}.
Only very recently larger datasets with accurate depth maps have been released~\cite{handa16}, or have been added to existing benchmarks~\cite{butler12}.
In our method we also make use of synthetically rendered data, but produce them in a much larger quantity.

For this purpose we implemented a ray-caster~\cite{appel68} that runs on the GPU and enables us to generate thousands of synthetic depth maps of high quality in a few minutes. 
For each image we randomly place between $24$ and $42$ rectangular cuboids and up to $3$ spheres in a predefined volume.
Further, we randomly scale and rotate each solid to achieve an infinitely number of possible constellations.
Then, we place a virtual camera at the origin of the coordinate system and cast a ray for each pixel of the camera image.
For each ray we compute the distance between the image plane and the closest surface it hits, or in the case it does not hit any surface, we return a maximum distance value for the background. 
In Figure~\ref{fig:train_data} we illustrate two random examples of the more than $40,000$ depth maps that we have generated with this method.

Given this generated depth maps as noise free ground-truth, we create the low-resolution depth maps $\smpli^\text{(lr)} =\, \downarrow_\scale \trgti$ for the network training by resampling the generated ground-truth depth maps $\trgti$ by the scale factor of $\scale$ that is used in the evaluation.
Depending on the dataset, we additionally add depth-dependent noise $\eta(\smpli^\text{(lr)})$ to the low-resolution depth map.
Finally, we upsample this low-resolution, probably noisy depth maps with bilinear interpolation to obtain our mid-level representation $\smpli =\, \uparrow_\scale\!(\smpli^\text{(lr)} + \eta(\smpli^\text{(lr)}))$.

\subsection{Clean Middlebury \& Laserscan}\label{sec:eval_mb_ls}

\begin{table}[tb]
  \centering
  \caption{
    Results on the noise-free Middlebury and Laserscan data.
    We report the error as root mean squared error (RMSE) in pixel disparity for the Middlebury data and in $mm$ for the Laserscan data, respectively.
    We highlight the best result in boldface and the second best in italic.
  }
  \tiny{
    \begin{tabular}{l r r r r r r r r r r r } \toprule
 & \multicolumn{4}{c}{$\times2$} & \multicolumn{4}{c}{$\times4$} & \multicolumn{3}{c}{$\times4$} \\ \cmidrule(lr){2-5}\cmidrule(lr){6-9}\cmidrule(lr){10-12}
 & Cones & Teddy & Tsukuba & Venus & Cones & Teddy & Tsukuba & Venus & Scan21 & Scan30 & Scan42\\ \midrule
NN & $4.3772$ & $3.2596$ & $9.7968$ & $2.1408$ & $6.1236$ & $4.5168$ & $13.3248$ & $2.9432$ & $0.0177$ & $0.0163$ & $0.0396$\\ 
Bicubic & $3.8392$ & $2.7668$ & $8.3648$ & $1.8192$ & $4.9544$ & $3.5744$ & $10.6960$ & $2.3504$ & $0.0132$ & $0.0125$ & $0.0326$\\ 
Diebel \& Thrun \cite{diebel05} & $2.9588$ & $2.1060$ & $6.4208$ & $1.3624$ & $4.5624$ & $3.2040$ & $8.7840$ & $1.9408$ & $-$ & $-$ & $-$\\ 
Ferstl \etal \cite{ferstl13}  & $2.8240$ & $2.1408$ & $7.0592$ & $1.2840$ & $3.6372$ & $2.5068$ & $10.0128$ & $1.4624$ & $-$ & $-$ & $-$\\ 
Zeyde \etal \cite{zeyde10} & $2.7680$ & $1.9616$ & $6.1936$ & $1.3200$ & $3.8468$ & $2.7812$ & $8.7632$ & $1.7592$ & $0.0100$ & $0.0093$ & $0.0246$\\ 
Timofte \etal \cite{timofte13} & $2.7872$ & $1.9816$ & $6.1280$ & $1.3328$ & $3.0256$ & $3.0256$ & $9.6304$ & $1.9616$ & $0.0106$ & $0.0101$ & $0.0264$\\ 
Aodha \etal \cite{aodha12} & $4.5076$ & $3.2988$ & $9.6192$ & $2.2088$ & $6.0168$ & $4.1036$ & $13.3328$ & $2.6920$ & $0.0175$ & $0.0170$ & $0.0452$\\ 
Horn\'{a}\v{c}ek \etal \cite{hornacek13} & $3.9744$ & $3.1640$ & $9.2832$ & $2.0592$ & $5.5944$ & $4.7828$ & $11.6352$ & $3.6008$ & $0.0205$ & $0.0179$ & $0.0299$\\ 
Ferstl \etal \cite{ferstl15} & $2.4988$ & $1.7588$ & $5.6064$ & $1.1464$ & $3.7336$ & $2.6680$ & $7.8416$ & $1.8096$ & $0.0085$ & $0.0083$ & $0.0190$\\  \midrule
CNN only & $1.0275$ & $\mathit{0.8201}$ & $\mathit{2.3610}$ & $\mathit{0.2266}$ & $3.0015$ & $1.5330$ & $\mathit{6.4361}$ & $0.4219$ & $\mathit{0.0083}$ & $\mathit{0.0082}$ & $\mathit{0.0120}$\\ 
CNN + ATGV-L2 & $\mathit{1.0145}$ & $0.8374$ & $\mathbf{2.3197}$ & $0.2720$ & $\mathit{2.9832}$ & $\mathit{1.5175}$ & $\mathbf{6.4223}$ & $\mathit{0.4124}$ & $0.0084$ & $0.0083$ & $0.0120$\\ 
ATGV-Net & $\mathbf{1.0021}$ & $\mathbf{0.8155}$ & $2.3846$ & $\mathbf{0.1991}$ & $\mathbf{2.9293}$ & $\mathbf{1.5029}$ & $6.6327$ & $\mathbf{0.3764}$ & $\mathbf{0.0081}$ & $\mathbf{0.0081}$ & $\mathbf{0.0117}$\\ 
\bottomrule\end{tabular}

  }
  \label{tab:results_mb2_laserscan}
\end{table}

In this first experiment we evaluate the performance of our proposed method on the images \emph{Teddy}, \emph{Cones}, \emph{Tsukuba} and \emph{Venus} of the Middlebury dataset as in~\cite{aodha12,ferstl15,hornacek13}.
The disparity is interpreted as depth and we test upsampling factors of $\times 2$ and $\times 4$. 
Additionally, we evaluate on the Laserscan dataset images \emph{Scan21}, \emph{Scan30} and \emph{Scan42} with an upsampling factor of $\times 4$ as in~\cite{aodha12,ferstl15}.
We compare our results to simple upsamling methods, such as nearest neighbor and bicubic upsampling, as well as to \sota depth upsampling methods that rely on an additional guidance image as input~\cite{diebel05,ferstl13}.
Further, we show the results of recent sparse coding based approaches for single image super-resolution~\cite{zeyde10,timofte13}, two approaches based on a Markov Random Field~\cite{aodha12,hornacek13} and a recent variational approach that uses sparse coding to estimate edge priors~\cite{ferstl15}.
To demonstrate the effect of our variational model on top of the deep network, we show the results of the high-resolution estimates of the network only (CNN only), the results, where we add the variational model, but without joint training (CNN + ATGV-L2), and the results after end-to-end training (\emph{ATGV-Net}).

\begin{figure}[tb]
  \centering
  \begin{minipage}{0.19\textwidth} \centering
    \includegraphics[width=\textwidth]{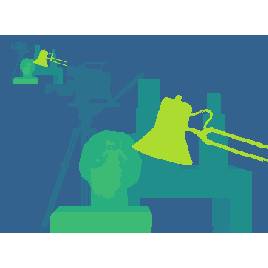}
    {\tiny (a) Input \& GT}
  \end{minipage}
  \begin{minipage}{0.19\textwidth} \centering
    \includegraphics[width=\textwidth]{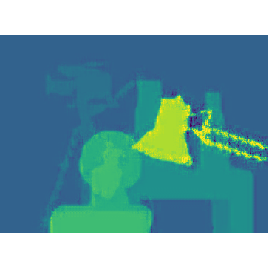}
    {\tiny (b) Timofte~\etal~\cite{timofte13}}
  \end{minipage}
  \begin{minipage}{0.19\textwidth} \centering
    \includegraphics[width=\textwidth]{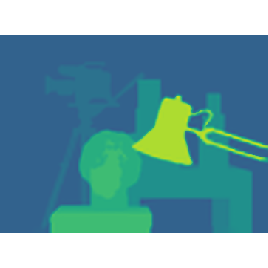}
    {\tiny (c) Ferstl~\etal~\cite{ferstl15}}
  \end{minipage}
  \begin{minipage}{0.19\textwidth} \centering
    \includegraphics[width=\textwidth]{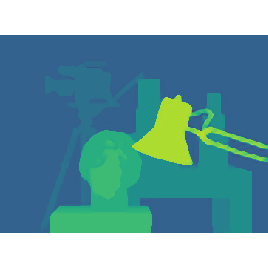}
    {\tiny (d) CNN only}
  \end{minipage}
  \begin{minipage}{0.19\textwidth} \centering
    \includegraphics[width=\textwidth]{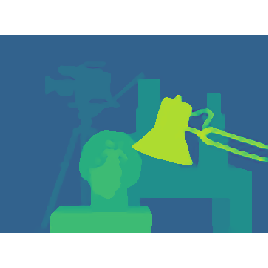}
    {\tiny (e) ATGV-Net}
  \end{minipage}
  \caption{
    Qualitative results for the noise-free Middlebury image \emph{Tsukuba}, $\scale = 4$.
    (a) depicts the ground-truth and the input data.
    (b) and (c) show the results of \sota methods.
    (d) and (e) present the results of the deep network only and our proposed model trained end-to-end.
    Best viewed magnified in the electronic version.
  }
  \label{fig:qual_results_mb2x4}
\end{figure}

The results in terms of the root mean squared error (RMSE) are summarized in Table~\ref{tab:results_mb2_laserscan}.\footnote{Note that we present our results over the full disparity range $[0, 255]$, as opposed to \eg~\cite{ferstl15}, where the disparities are scaled to a narrower range.}
We can clearly see that the deep network already achieves a significant performance improvement compared to the other methods on both datasets and upsampling factors.
Interestingly, we obtain even better results as the methods \cite{diebel05,ferstl13} that utilize an additional guidance image for the upsampling.
This is especially pronounced in test samples with structures that are well simulated in the training data, such as \emph{Venus}.
Further, the variational model on top of the network slightly increases the performance and training the whole model end-to-end gives the overall best results.
One exception is the \emph{Tsukuba} sample, where the results get slightly worse after end-to-end training.
An explanation might be that fine, elongated structures, \eg near at the lamp of \emph{Tsukuba}, are not well represented in the training data.
In the qualitative results, see Figure~\ref{fig:qual_results_mb2x4}, we can further observe that the deep network with $10$ layers achieves already very good results with sharper depth discontinuities compared to other methods.
However, the improvement of the variational model on top of the deep network is hardly visible.
This becomes more apparent in the next experiment.

\subsection{Noisy Middlebury}\label{sec:eval_noisy_mb}

\begin{table}[tb]
  \centering
  \caption{
    Results on noisy Middlebury data.
    We report the error as RMSE in pixel disparity and highlight the best result in boldface and the second best in italic.
  }
  \tiny{
    \begin{tabular}{l r r r r r r } \toprule
 & \multicolumn{3}{c}{$\times2$} & \multicolumn{3}{c}{$\times4$} \\ \cmidrule(lr){2-4}\cmidrule(lr){5-7}
 & Art & Books & Moebius & Art & Books & Moebius\\ \midrule
NN & $6.55$ & $6.16$ & $6.59$ & $7.48$ & $6.31$ & $6.78$\\ 
Bilinear & $4.58$ & $3.95$ & $4.20$ & $5.62$ & $4.31$ & $4.56$\\ 
Yang \etal \cite{yang07} & $3.01$ & $1.87$ & $1.92$ & $4.02$ & $2.38$ & $2.42$\\ 
He \etal \cite{he10} & $3.55$ & $2.37$ & $2.48$ & $4.41$ & $2.74$ & $2.83$\\ 
Diebel \& Thrun \cite{diebel05} & $3.49$ & $2.06$ & $2.13$ & $4.51$ & $3.00$ & $3.11$\\ 
Chan \etal \cite{chan08} & $3.44$ & $2.09$ & $2.08$ & $4.46$ & $2.77$ & $2.76$\\ 
Park \etal \cite{park11} & $3.76$ & $1.95$ & $1.96$ & $4.56$ & $2.61$ & $2.51$\\ 
  Ferstl \etal \cite{ferstl13}  & $3.19$ & $1.52$ & $1.47$ & $4.06$ & $\mathit{2.21}$ & $\mathit{2.03}$\\  \midrule
CNN only & $2.02$ & $1.27$ & $1.50$ & $3.55$ & $2.41$ & $2.68$\\ 
  CNN + ATGV-L2 & $\mathit{1.93}$ & $\mathit{1.14}$ & $\mathit{1.37}$ & $\mathit{3.40}$ & $2.24$ & $2.51$\\ 
ATGV-Net & $\mathbf{1.84}$ & $\mathbf{1.13}$ & $\mathbf{1.24}$ & $\mathbf{2.98}$ & $\mathbf{1.72}$ & $\mathbf{1.95}$\\ 
\bottomrule\end{tabular}

  }
  \label{tab:results_noisy_mb}
\end{table}

\begin{figure}[tb]
  \centering
  \begin{minipage}{0.19\textwidth} \centering
    \includegraphics[width=\textwidth]{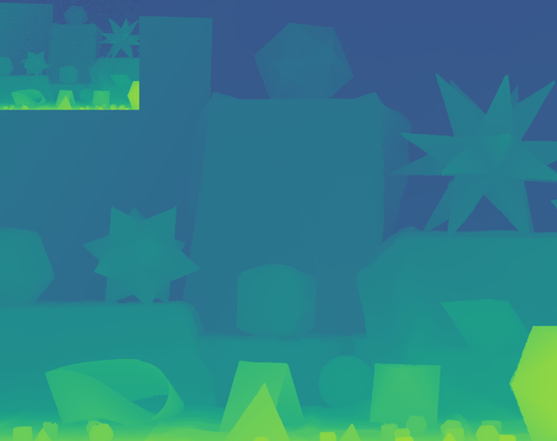}
    {\tiny (a) Input \& GT}
  \end{minipage}
  \begin{minipage}{0.19\textwidth} \centering
    \includegraphics[width=\textwidth]{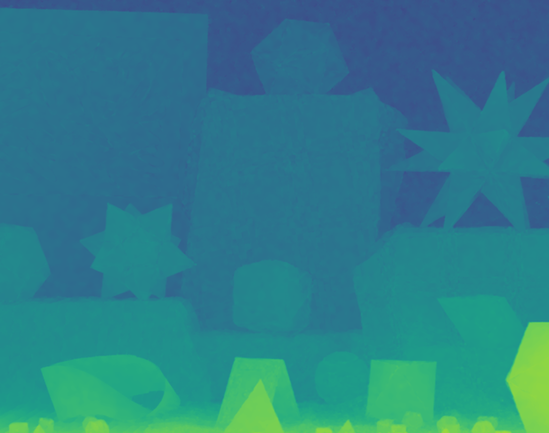}
    {\tiny (b) He~\etal~\cite{he10}}
  \end{minipage}
  \begin{minipage}{0.19\textwidth} \centering
    \includegraphics[width=\textwidth]{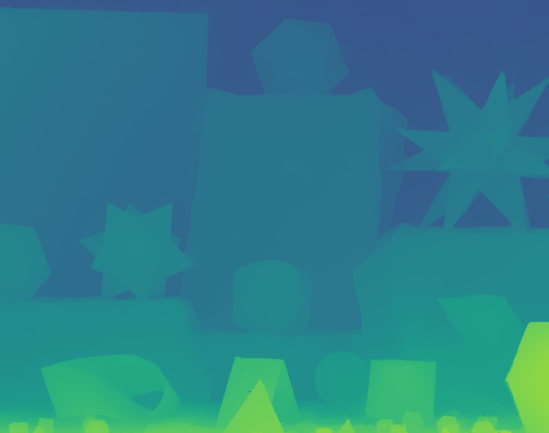}
    {\tiny (c) Ferstl~\etal~\cite{ferstl13}}
  \end{minipage}
  \begin{minipage}{0.19\textwidth} \centering
    \includegraphics[width=\textwidth]{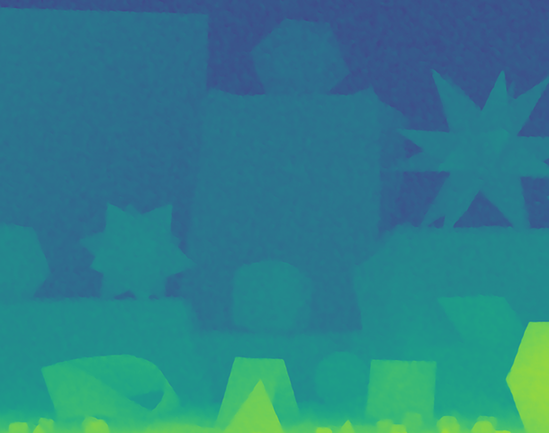}
    {\tiny (d) CNN only}
  \end{minipage}
  \begin{minipage}{0.19\textwidth} \centering
    \includegraphics[width=\textwidth]{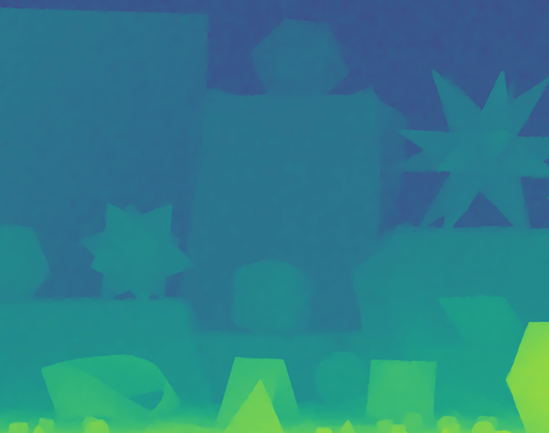}
    {\tiny (e) ATGV-Net}
  \end{minipage}
  \caption{
    Qualitative results for the noisy Middlebury image \emph{Moebius}, $\scale = 4$.
    (a) depicts the ground-truth and the input data.
    (b) and (c) show the results of \sota methods.
    (d) and (e) present the results of the deep network only and our proposed model trained end-to-end.
    Best viewed magnified in the electronic version.
  }
  \label{fig:qual_results_nmbx4}
\end{figure}

In this experiment we evaluate our method on the Middlebury disparity maps \emph{Art}, \emph{Books} and \emph{Moebius} with added depth dependent Gaussian noise to simulate the acquisition process of a Time-of-Flight sensor, as proposed by Park~\etal~\cite{park11}.
Therefore, we add to our low-resolution synthetic training data $\smpli^\text{(lr)}$ depth dependent Gaussian noise of the form $\eta(x) = \mathcal{N}(0, \sigma \smpli^\text{(lr)}(x)^{-1})$, with $\sigma = 651$.
Exemplar training images are depicted in Figure~\ref{fig:train_data}.
We report quantitative results in Table~\ref{tab:results_noisy_mb} and visualize qualitative results in Figure~\ref{fig:qual_results_nmbx4}.

We again compare our method to simple upsampling methods, such as nearest neighbor and bilinear interpolation.
We compare our proposed method to other approaches that utilize an additional intensity image as guidance.
Those methods include the Markov Random Field based approach in~\cite{diebel05}, the bilateral filtering with cost volume in~\cite{yang07}, the guided image filter in~\cite{he10}, the noise-aware bilateral filter in~\cite{chan08}, the non-local means filter in~\cite{park11} and the variational model in~\cite{ferstl13}.
To evaluate the influence of the variational model on top of the deep network, we report the results of the network only (CNN only), results with the variational model on top of the network, but without joint training (CNN + ATGV-L2), and the results after end-to-end training (ATGV-Net).

From the quantitative results in Table~\ref{tab:results_noisy_mb} we observe that the \emph{CNN only} already performs better than \sota methods that utilize an additional guidance input for most images and upsampling factors.
Further, the variational model on top of the deep network slightly improves the results, but end-to-end training of the whole model results in significant improvement.
This improvement of \emph{ATGV-Net} over the network only is also apparent in the qualitative results (Figure~\ref{fig:qual_results_nmbx4}).
We observe less noise in homogeneous areas in the \emph{ATGV-Net} estimates, especially in the background, compared to the \emph{CNN only} estimates.
The results of \cite{ferstl13} look also very sharp, but produce errors near depth discontinuities and in-between fine structures.
In contrast, our method preserves those finer structures.
We refer to the supplemental material for additional qualitative results, as well as quantitative results in terms of mean absolute error (MAE).

\subsection{ToFMark}\label{sec:eval_tofmark}

In our final experiment we evaluate our method on the challenging ToFMark dataset~\cite{ferstl13}.
This dataset consists of three time-of-flight (ToF) depth maps of three different scenes.
For each scene there exists an accurate high-resolution structured-light scan as ground-truth.
The ToF depth maps have a resolution of $120 \times 160$ pixel and the target resolution, given by the guidance intensity image (that we do not use in our method) is $610 \times 810$ pixel.
This corresponds to an upsampling factor of approximately $\scale = 5$.
As the target high-resolution depth-map is given in the camera coordinate system of the structured light scanner, we prepare our training data accordingly. 
We project our high-resolution synthetic training depth maps to the ToF coordinate system using the provided projection matrix.
In the low-resolution depth maps we add depth dependent noise and back project the remaining points to the target camera coordinate system.
This yields a very sparse depth map that we subsequently inpaint with bilinear interpolation to obtain our final mid-resolution training inputs.

We compare our results to simple nearest neighbour and bilinear interpolation, and three \sota depth map super-resolution methods that utilize an additional guidance image as input.
The quantitative results are shown in Table~\ref{tab:results_tofmark} as RMSE in $mm$.
Please see the supplemental material for qualitative results.
Even on this difficult dataset we are at least on par with \sota methods that utilize an additional intensity image as guidance input.

\begin{table}[tb]
  \centering
  \begin{minipage}[c]{0.35\textwidth}
    \tiny{
      \begin{tabular}{l r r r } \toprule
 & Books & Devil & Shark\\ \midrule
NN & $30.46$ & $27.53$ & $38.21$\\ 
Bilinear & $29.11$ & $25.34$ & $36.34$\\ 
Kopf \etal \cite{kopf07} & $27.82$ & $24.30$ & $34.79$\\ 
He \etal \cite{he10} & $27.11$ & $23.45$ & $33.26$\\ 
Ferstl \etal \cite{ferstl13}  & $\mathbf{24.00}$ & $\mathit{23.19}$ & $\mathit{29.89}$\\  \midrule
ATGV-Net & $\mathit{24.67}$ & $\mathbf{21.74}$ & $\mathbf{28.51}$\\ 
\bottomrule\end{tabular}

    }
  \end{minipage}\hfill
  \begin{minipage}[c]{0.64\textwidth}
    \caption{
      Results on real Time-of-Flight data from the ToFMark benchmark dataset.
      We report the error as RMSE in $mm$ and highlight the best result in boldface and the second best in italic.
    }
    \label{tab:results_tofmark}
  \end{minipage}
\end{table}

\section{Conclusion}
We presented combination of a deep convolutional network with a variational model for single depth map super-resolution.
We designed the convolutional to compute the high-resolution depth map, as well as the depth discontinuities.
The network output was utilized in our variational model to further refine the result.
By unrolling the optimization procedure of the variational model, we were able to optimize the joint model end-to-end, which lead to improved accuracy.
Further, we demonstrated the feasibility to train our method on a massive amount of synthetic generated depth data and obtain \sota results on four different benchmarks.
Our model is especially useful if the low-resolution depth map contains noise, which is the case for most consumer depth sensors. 
In future work we plan to extend our model to depth data that contain larger areas of missing pixels, \eg from structured light sensors.
This is straight-forward by setting $\wgts_\lambda = 0$ for areas where depth measurements are missing.

{\footnotesize\textbf{Acknowledgment:} This work was supported by \emph{Infineon Technologies Austria AG} and the Austrian Research Promotion Agency under the \emph{FIT-IT Bridge} program, project \#838513 (TOFUSION).}

\clearpage

\bibliographystyle{splncs03}
\bibliography{index}
\clearpage

\section{Supplemental Material}

\subsection{Overview}
\begin{figure}[h]
  \centering
  \begin{sideways}
    \includegraphics[width=0.58\textheight]{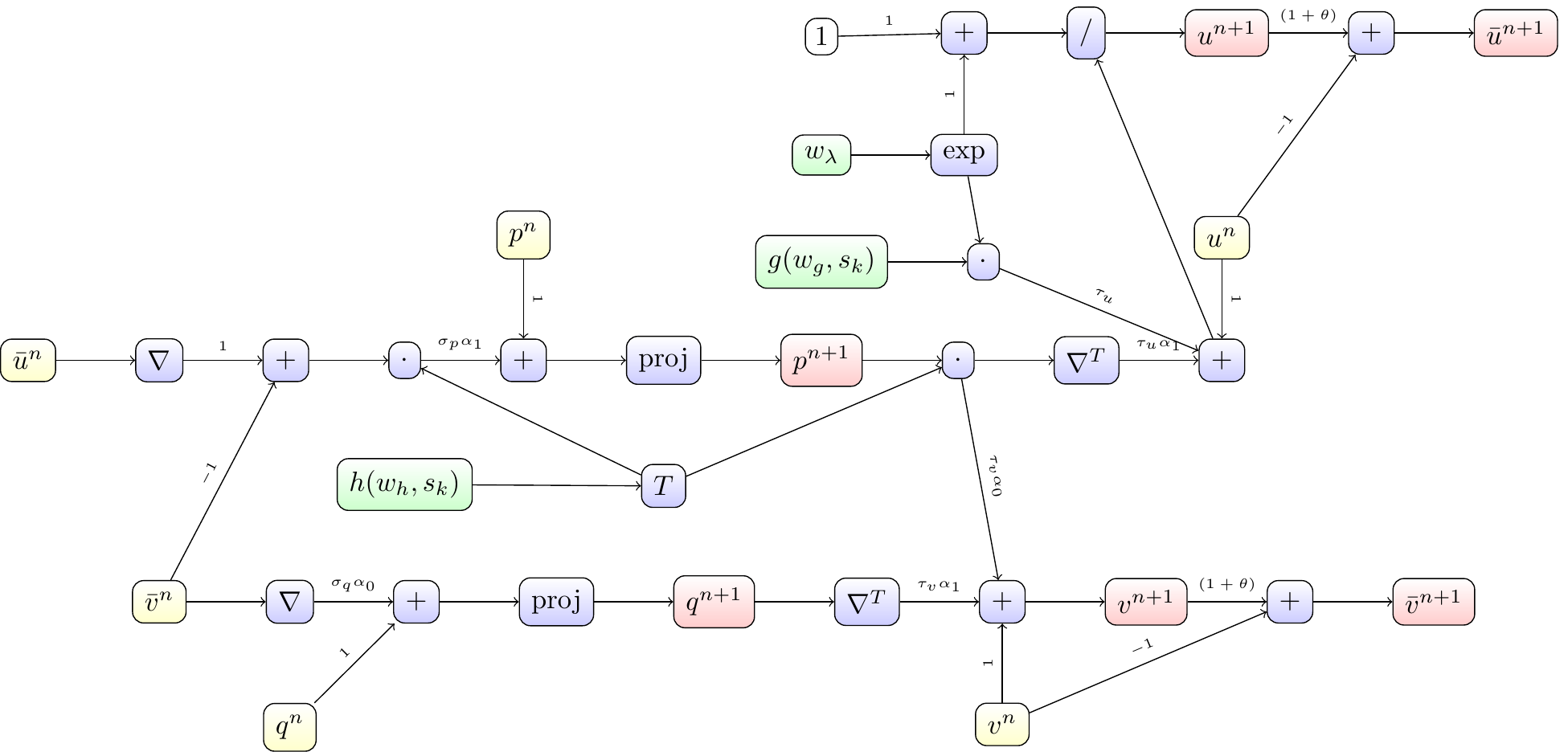}
  \end{sideways}
  \caption{
    A single iteration of the first-order primal-dual algorithm by \cite{chambolle11} in terms of operations in a deep network.
    The green nodes represent inputs form the parametrization $f(\wgts, \smpli) = [g(\wgts_g, \smpli), \wgts_\lambda, h(\wgts_h, \smpli)]^T$, \ie our deep network.
    The values from the previous iteration and the outputs of this iteration are colored in yellow and red, respectively.
    The operations that perform the computations of the iteration are depicted as blue rectangles, and correspond to layers in our network.
    The values on the edges of the graph indicate the additional weights in the pointwise additions.
    Best viewed in color.
  }
  \label{fig:computation_graph}
\end{figure}

The supplementary material of our ECCV 2016 submission provides a graphical representation of one primal-dual iteration realized by layers of a deep network, as well as additional quantitative and qualitative results. 
In Figure~\ref{fig:computation_graph} of Section~\ref{sec:computation_graph}, we have a feed-forward graph that details all steps of one primal-dual iteration.
Nodes in green represent inputs from the parametrization $f(\wgts, \smpli) = [g(\wgts_g, \smpli), \wgts_\lambda, h(\wgts_h, \smpli)]^T$, \ie our deep network.
The nodes in yellow are inputs from a previous iteration of the primal-dual algorithm.
Blue nodes perform actual computations and are realized by network layers and red nodes are the outputs of the iteration.

\begin{figure}[h]
  \centering
  \begin{minipage}{0.24\textwidth} \centering
    \includegraphics[width=\textwidth]{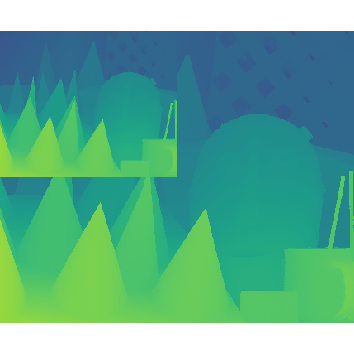}
    \mbox{\phantom{\includegraphics[width=\textwidth]{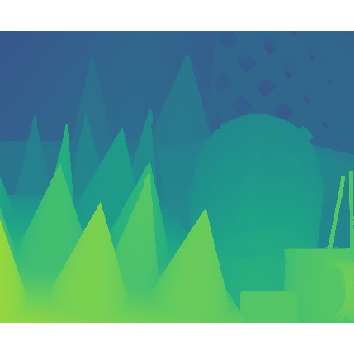}}}
    {\tiny (a) Input \& GT}
  \end{minipage}
  \begin{minipage}{0.24\textwidth} \centering
    \includegraphics[width=\textwidth]{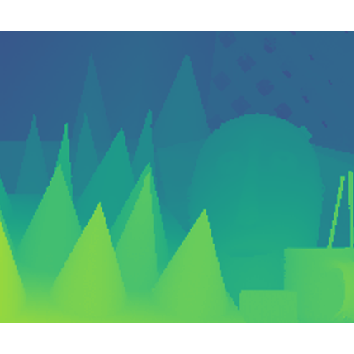}
    \includegraphics[width=\textwidth]{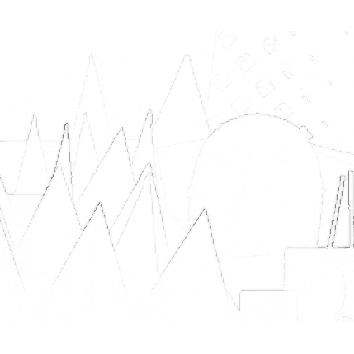}
    {\tiny (b) Bicubic}
  \end{minipage}
  \begin{minipage}{0.24\textwidth} \centering
    \includegraphics[width=\textwidth]{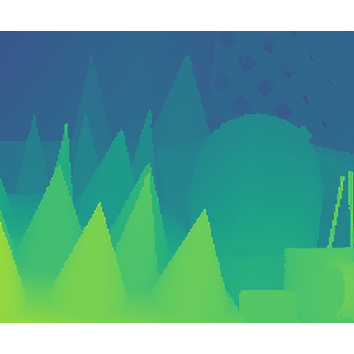}
    \includegraphics[width=\textwidth]{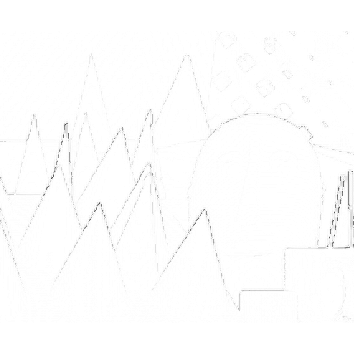}
    {\tiny (c) Zeyde~\etal~\cite{zeyde10}}
  \end{minipage}
  \begin{minipage}{0.24\textwidth} \centering
    \includegraphics[width=\textwidth]{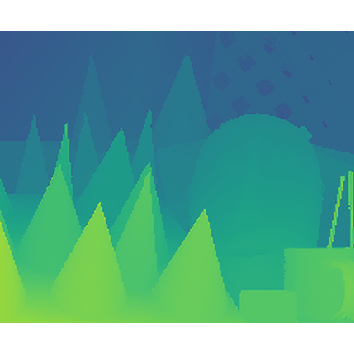}
    \includegraphics[width=\textwidth]{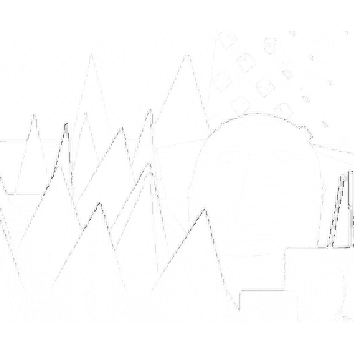}
    {\tiny (d) Timofte~\etal~\cite{timofte13}}
  \end{minipage}
  \begin{minipage}{0.24\textwidth} \centering
    \includegraphics[width=\textwidth]{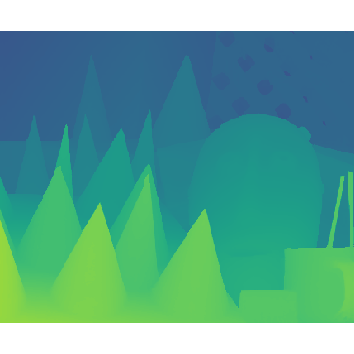}
    \includegraphics[width=\textwidth]{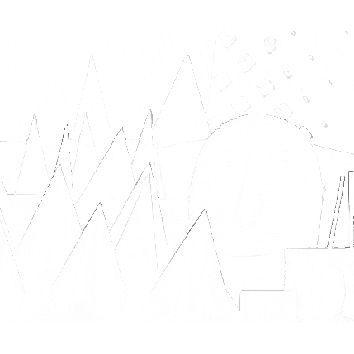}
    {\tiny (e) Horn\'{a}\v{c}ek~\etal~\cite{hornacek13}}
  \end{minipage}
  \begin{minipage}{0.24\textwidth} \centering
    \includegraphics[width=\textwidth]{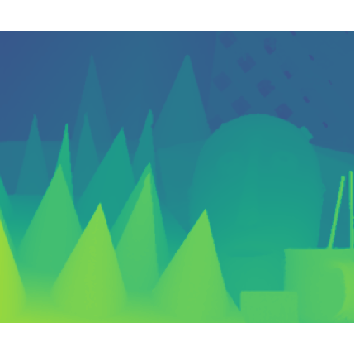}
    \includegraphics[width=\textwidth]{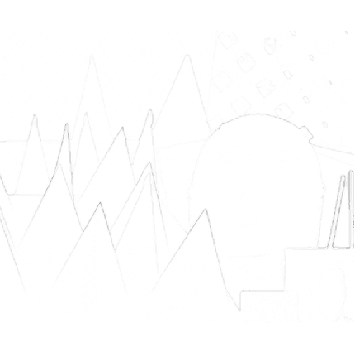}
    {\tiny (f) Ferstl~\etal~\cite{ferstl15}}
  \end{minipage}
  \begin{minipage}{0.24\textwidth} \centering
    \includegraphics[width=\textwidth]{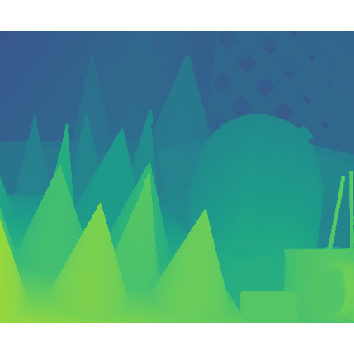}
    \includegraphics[width=\textwidth]{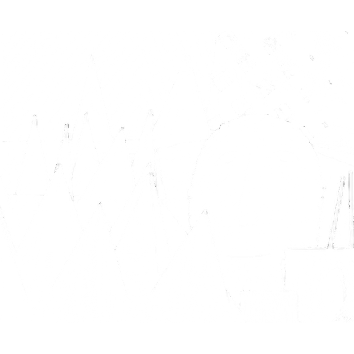}
    {\tiny (g) CNN only}
  \end{minipage}
  \begin{minipage}{0.24\textwidth} \centering
    \includegraphics[width=\textwidth]{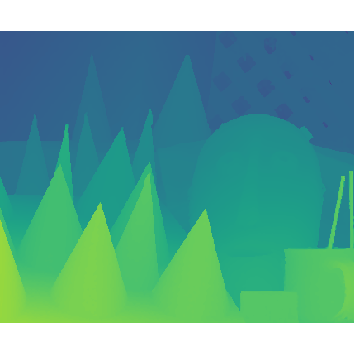}
    \includegraphics[width=\textwidth]{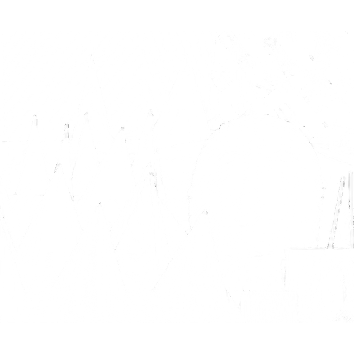}
    {\tiny (h) ATGV-Net}
  \end{minipage}
  \caption{
    Qualitative results for the noise-free Middlebury dataset sample \emph{Cones} and an upsampling factor $\scale = \times 2$.
    In (a) we show the ground-truth high-resolution depth map along with the low-resolution input in the top left corner, preserving the relative resolution.
    In (b) to (f) we visualize the results of bicubic upsampling and \sota approaches with the corresponding error images.
    In (g) we show the result of our network only, and in (h) we depict the result of our proposed \emph{ATGV-Net}.
    Best viewed magnified in the electronic version.
  }
  \label{fig:qual_results_mb2x2_00}
\end{figure}

\begin{figure}[b]
  \centering
  \begin{minipage}{0.24\textwidth} \centering
    \includegraphics[width=\textwidth]{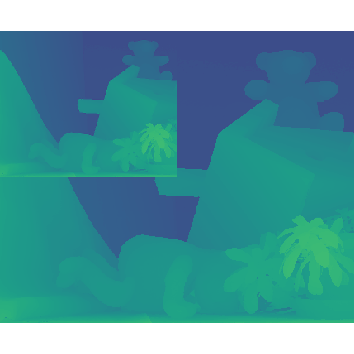}
    \mbox{\phantom{\includegraphics[width=\textwidth]{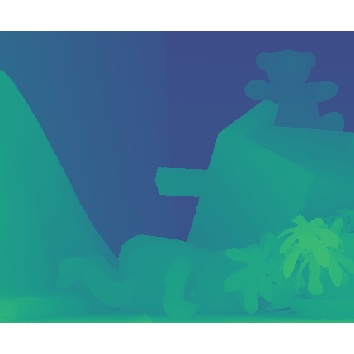}}}
    {\tiny (a) Input \& GT}
  \end{minipage}
  \begin{minipage}{0.24\textwidth} \centering
    \includegraphics[width=\textwidth]{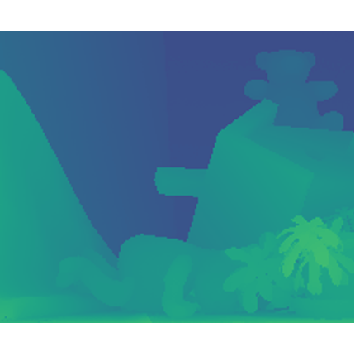}
    \includegraphics[width=\textwidth]{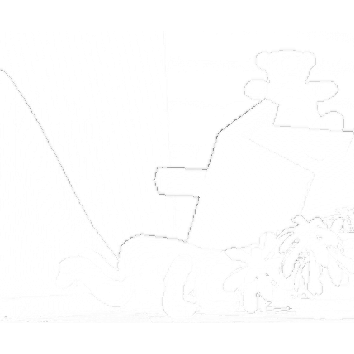}
    {\tiny (b) Bicubic}
  \end{minipage}
  \begin{minipage}{0.24\textwidth} \centering
    \includegraphics[width=\textwidth]{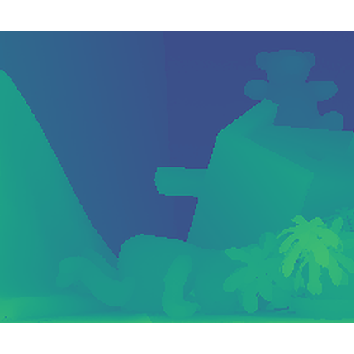}
    \includegraphics[width=\textwidth]{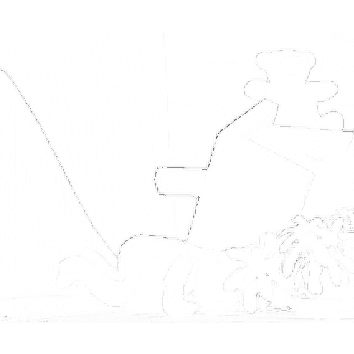}
    {\tiny (c) Zeyde~\etal~\cite{zeyde10}}
  \end{minipage}
  \begin{minipage}{0.24\textwidth} \centering
    \includegraphics[width=\textwidth]{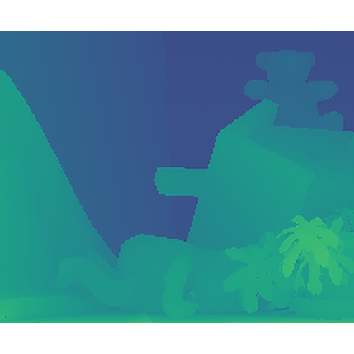}
    \includegraphics[width=\textwidth]{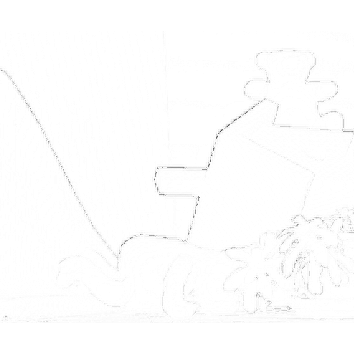}
    {\tiny (d) Timofte~\etal~\cite{timofte13}}
  \end{minipage}
  \begin{minipage}{0.24\textwidth} \centering
    \includegraphics[width=\textwidth]{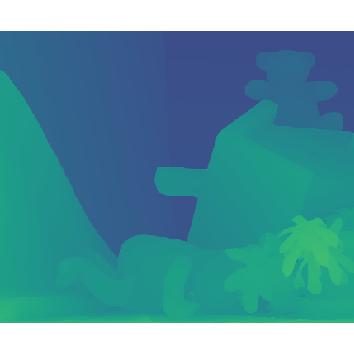}
    \includegraphics[width=\textwidth]{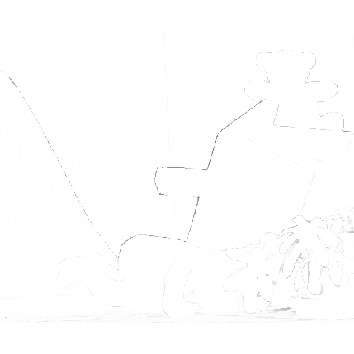}
    {\tiny (e) Horn\'{a}\v{c}ek~\etal~\cite{hornacek13}}
  \end{minipage}
  \begin{minipage}{0.24\textwidth} \centering
    \includegraphics[width=\textwidth]{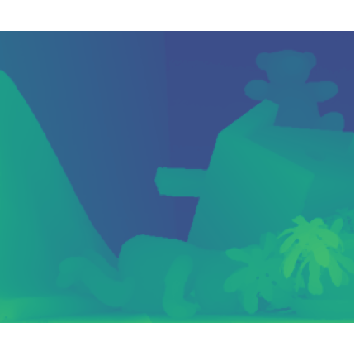}
    \includegraphics[width=\textwidth]{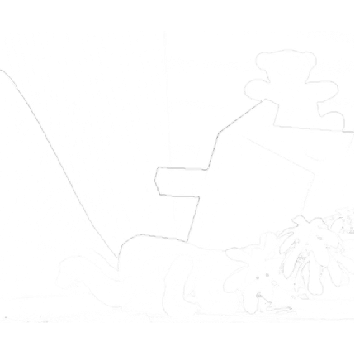}
    {\tiny (f) Ferstl~\etal~\cite{ferstl15}}
  \end{minipage}
  \begin{minipage}{0.24\textwidth} \centering
    \includegraphics[width=\textwidth]{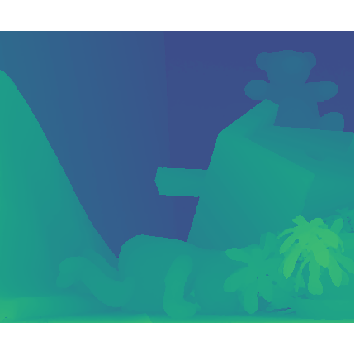}
    \includegraphics[width=\textwidth]{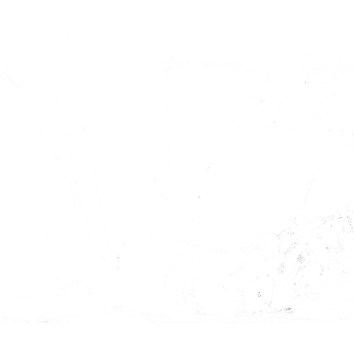}
    {\tiny (g) CNN only}
  \end{minipage}
  \begin{minipage}{0.24\textwidth} \centering
    \includegraphics[width=\textwidth]{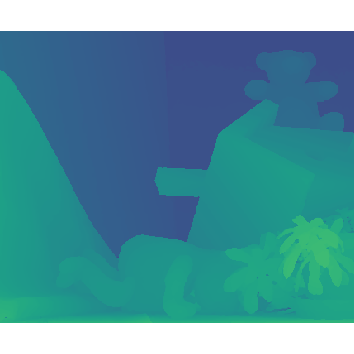}
    \includegraphics[width=\textwidth]{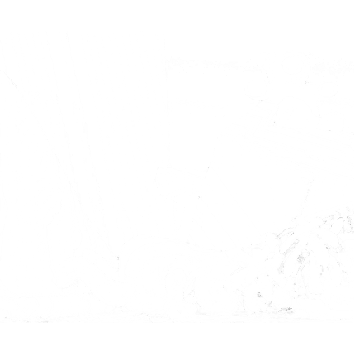}
    {\tiny (h) ATGV-Net}
  \end{minipage}
  
  \caption{
    Qualitative results for the noise-free Middlebury dataset sample \emph{Teddy} and an upsampling factor $\scale = \times 2$.
    In (a) we show the ground-truth high-resolution depth map along with the low-resolution input in the top left corner, preserving the relative resolution.
    In (b) to (f) we visualize the results of bicubic upsampling and \sota approaches with the corresponding error images.
    In (g) we show the result of our network only, and in (h) we depict the result of our proposed \emph{ATGV-Net}.
    Best viewed magnified in the electronic version.
  }
  \label{fig:qual_results_mb2x2_01}
\end{figure}

\begin{figure}[b]
  \centering
  \begin{minipage}{0.24\textwidth} \centering
    \includegraphics[width=\textwidth]{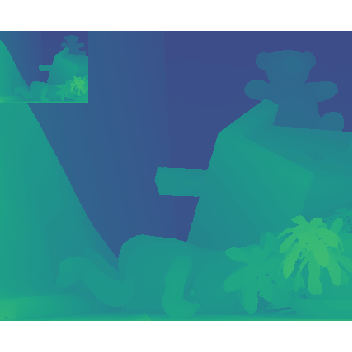}
    \mbox{\phantom{\includegraphics[width=\textwidth]{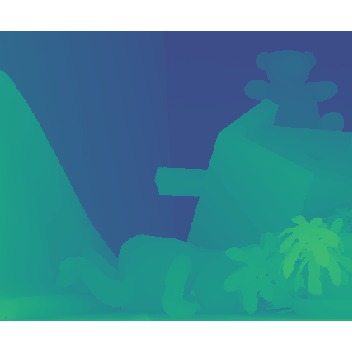}}}
    {\tiny (a) Input \& GT}
  \end{minipage}
  \begin{minipage}{0.24\textwidth} \centering
    \includegraphics[width=\textwidth]{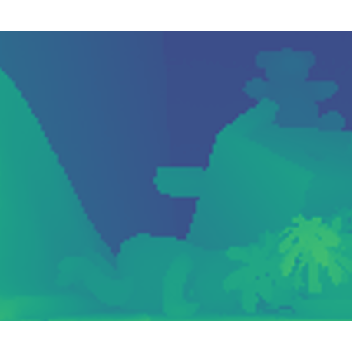}
    \includegraphics[width=\textwidth]{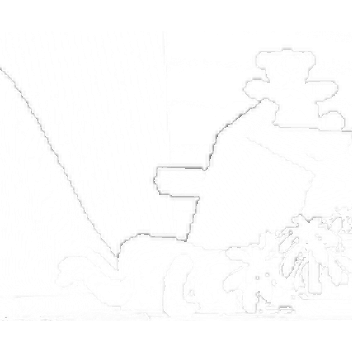}
    {\tiny (b) Bicubic}
  \end{minipage}
  \begin{minipage}{0.24\textwidth} \centering
    \includegraphics[width=\textwidth]{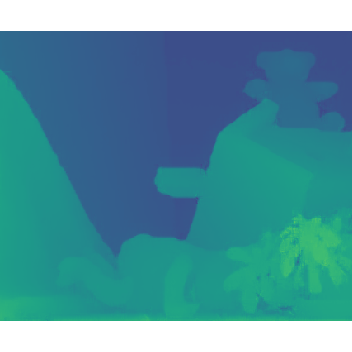}
    \includegraphics[width=\textwidth]{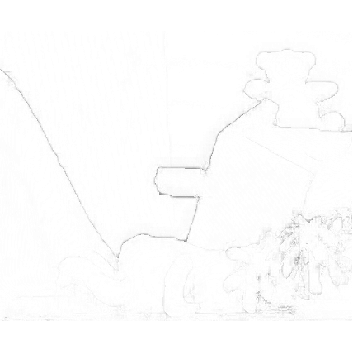}
    {\tiny (c) Zeyde~\etal~\cite{zeyde10}}
  \end{minipage}
  \begin{minipage}{0.24\textwidth} \centering
    \includegraphics[width=\textwidth]{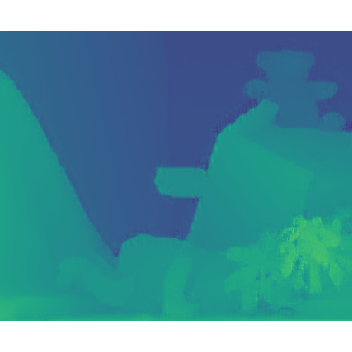}
    \includegraphics[width=\textwidth]{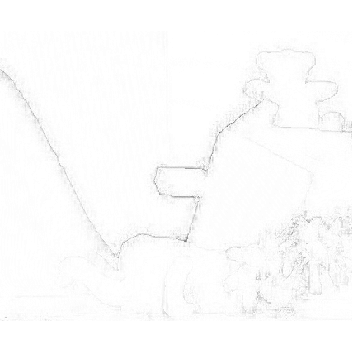}
    {\tiny (d) Timofte~\etal~\cite{timofte13}}
  \end{minipage}
  \begin{minipage}{0.24\textwidth} \centering
    \includegraphics[width=\textwidth]{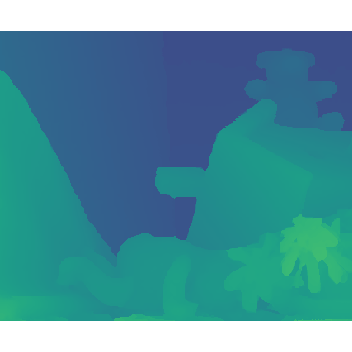}
    \includegraphics[width=\textwidth]{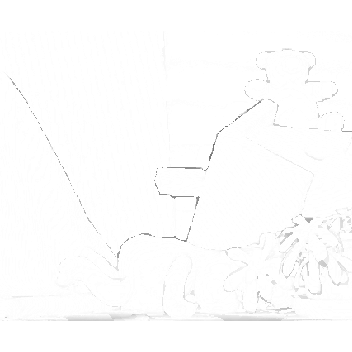}
    {\tiny (e) Horn\'{a}\v{c}ek~\etal~\cite{hornacek13}}
  \end{minipage}
  \begin{minipage}{0.24\textwidth} \centering
    \includegraphics[width=\textwidth]{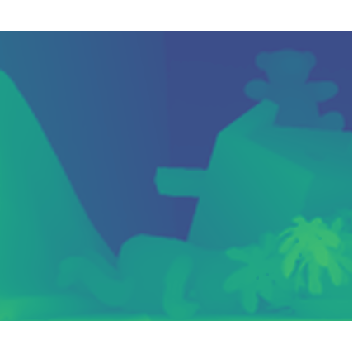}
    \includegraphics[width=\textwidth]{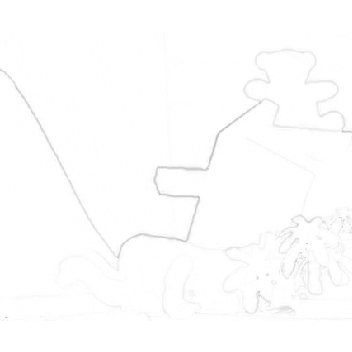}
    {\tiny (f) Ferstl~\etal~\cite{ferstl15}}
  \end{minipage}
  \begin{minipage}{0.24\textwidth} \centering
    \includegraphics[width=\textwidth]{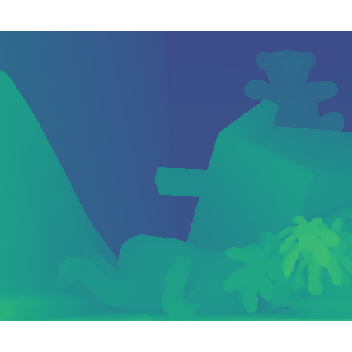}
    \includegraphics[width=\textwidth]{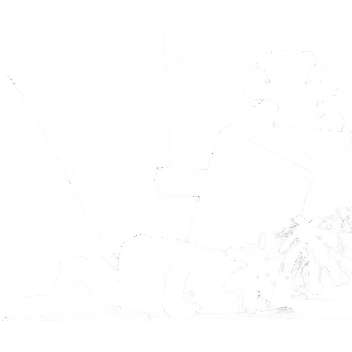}
    {\tiny (g) CNN only}
  \end{minipage}
  \begin{minipage}{0.24\textwidth} \centering
    \includegraphics[width=\textwidth]{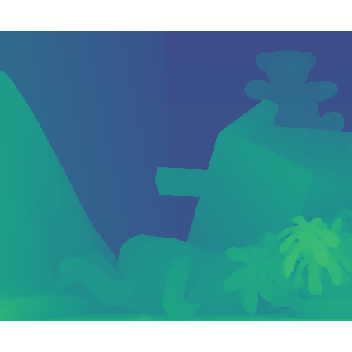}
    \includegraphics[width=\textwidth]{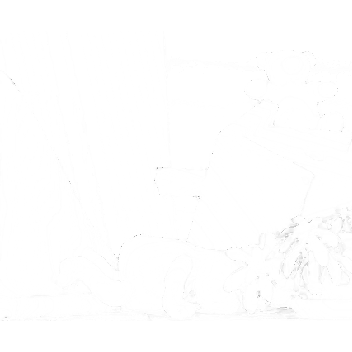}
    {\tiny (h) ATGV-Net}
  \end{minipage}
  
  \caption{
    Qualitative results for the noise-free Middlebury dataset sample \emph{Teddy} and an upsampling factor $\scale = \times 4$.
    In (a) we show the ground-truth high-resolution depth map along with the low-resolution input in the top left corner, preserving the relative resolution.
    In (b) to (f) we visualize the results of bicubic upsampling and \sota approaches with the corresponding error images.
    In (g) we show the result of our network only, and in (h) we depict the result of our proposed \emph{ATGV-Net}.
    Best viewed magnified in the electronic version.
  }
  \label{fig:qual_results_mb2x4_01}
\end{figure}

\begin{figure}[b]
  \centering
  \begin{minipage}{0.24\textwidth} \centering
    \includegraphics[width=\textwidth]{results/mb2x4/ta_in_depth_02.png}
    \mbox{\phantom{\includegraphics[width=\textwidth]{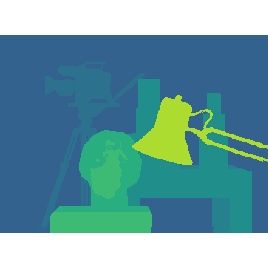}}}
    {\tiny (a) Input \& GT}
  \end{minipage}
  \begin{minipage}{0.24\textwidth} \centering
    \includegraphics[width=\textwidth]{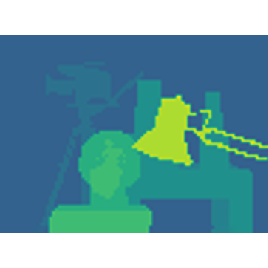}
    \includegraphics[width=\textwidth]{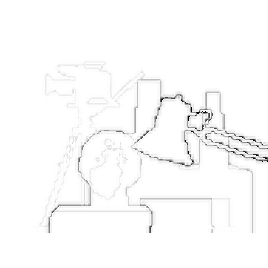}
    {\tiny (b) Bicubic}
  \end{minipage}
  \begin{minipage}{0.24\textwidth} \centering
    \includegraphics[width=\textwidth]{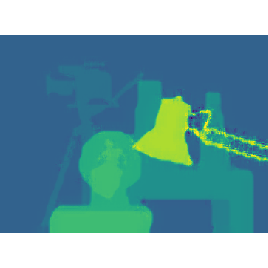}
    \includegraphics[width=\textwidth]{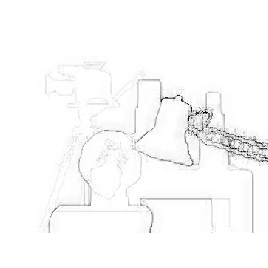}
    {\tiny (c) Zeyde~\etal~\cite{zeyde10}}
  \end{minipage}
  \begin{minipage}{0.24\textwidth} \centering
    \includegraphics[width=\textwidth]{results/mb2x4/timofte13_depth_02.png}
    \includegraphics[width=\textwidth]{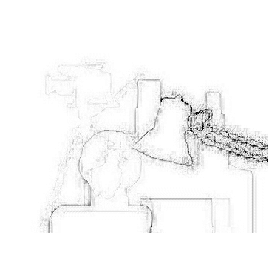}
    {\tiny (d) Timofte~\etal~\cite{timofte13}}
  \end{minipage}
  \begin{minipage}{0.24\textwidth} \centering
    \includegraphics[width=\textwidth]{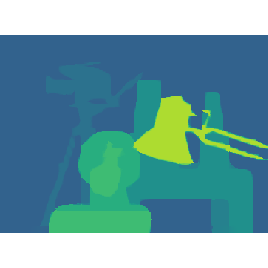}
    \includegraphics[width=\textwidth]{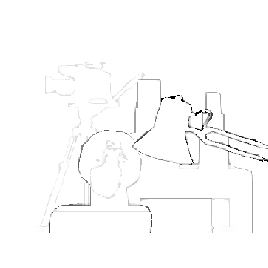}
    {\tiny (e) Horn\'{a}\v{c}ek~\etal~\cite{hornacek13}}
  \end{minipage}
  \begin{minipage}{0.24\textwidth} \centering
    \includegraphics[width=\textwidth]{results/mb2x4/ferstl15_depth_02.png}
    \includegraphics[width=\textwidth]{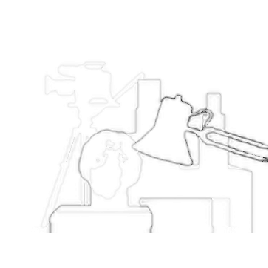}
    {\tiny (f) Ferstl~\etal~\cite{ferstl15}}
  \end{minipage}
  \begin{minipage}{0.24\textwidth} \centering
    \includegraphics[width=\textwidth]{results/mb2x4/srcnn10_depth_02.png}
    \includegraphics[width=\textwidth]{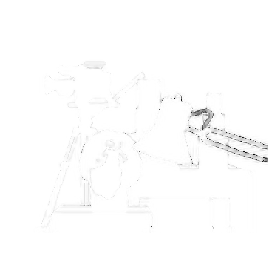}
    {\tiny (g) CNN only}
  \end{minipage}
  \begin{minipage}{0.24\textwidth} \centering
    \includegraphics[width=\textwidth]{results/mb2x4/atgvl2_srcnn10_depth_02.png}
    \includegraphics[width=\textwidth]{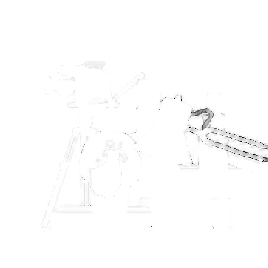}
    {\tiny (h) ATGV-Net}
  \end{minipage}
  
  \caption{
    Qualitative results for the noise-free Middlebury dataset sample \emph{Tsukuba} and an upsampling factor $\scale = \times 4$.
    In (a) we show the ground-truth high-resolution depth map along with the low-resolution input in the top left corner, preserving the relative resolution.
    In (b) to (f) we visualize the results of bicubic upsampling and \sota approaches with the corresponding error images.
    In (g) we show the result of our network only, and in (h) we depict the result of our proposed \emph{ATGV-Net}.
    Best viewed magnified in the electronic version.
  }
  \label{fig:qual_results_mb2x4_02}
\end{figure}

\begin{figure}[b]
  \centering
  \begin{minipage}{0.24\textwidth} \centering
    \includegraphics[width=\textwidth]{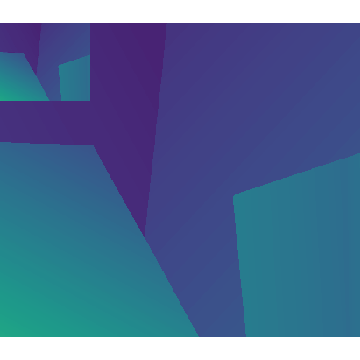}
    \mbox{\phantom{\includegraphics[width=\textwidth]{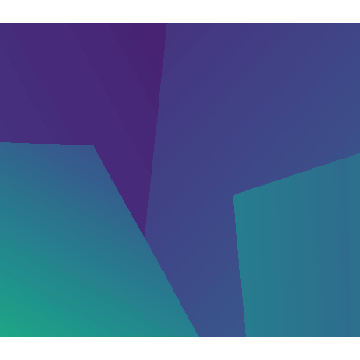}}}
    {\tiny (a) Input \& GT}
  \end{minipage}
  \begin{minipage}{0.24\textwidth} \centering
    \includegraphics[width=\textwidth]{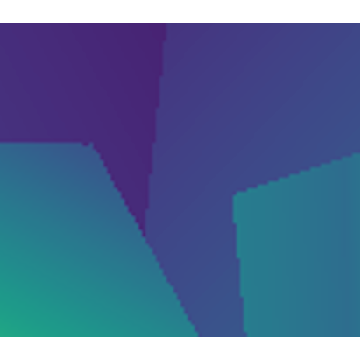}
    \includegraphics[width=\textwidth]{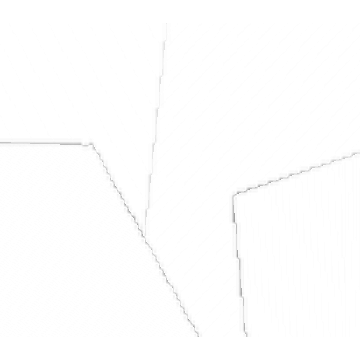}
    {\tiny (b) Bicubic}
  \end{minipage}
  \begin{minipage}{0.24\textwidth} \centering
    \includegraphics[width=\textwidth]{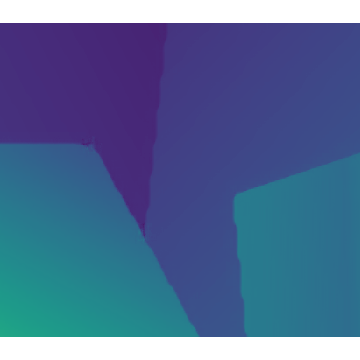}
    \includegraphics[width=\textwidth]{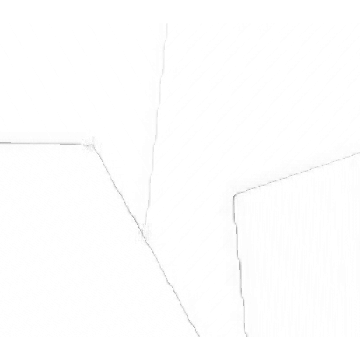}
    {\tiny (c) Zeyde~\etal~\cite{zeyde10}}
  \end{minipage}
  \begin{minipage}{0.24\textwidth} \centering
    \includegraphics[width=\textwidth]{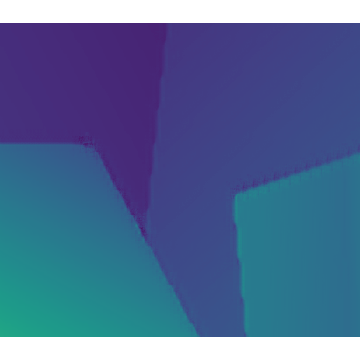}
    \includegraphics[width=\textwidth]{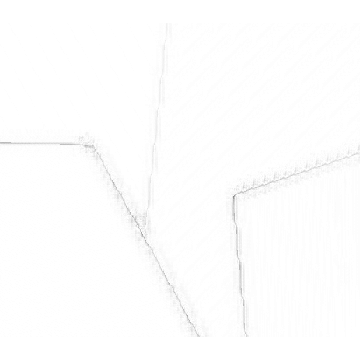}
    {\tiny (d) Timofte~\etal~\cite{timofte13}}
  \end{minipage}
  \begin{minipage}{0.24\textwidth} \centering
    \includegraphics[width=\textwidth]{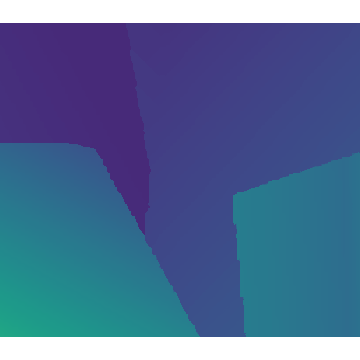}
    \includegraphics[width=\textwidth]{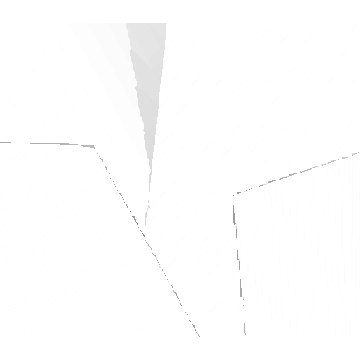}
    {\tiny (e) Horn\'{a}\v{c}ek~\etal~\cite{hornacek13}}
  \end{minipage}
  \begin{minipage}{0.24\textwidth} \centering
    \includegraphics[width=\textwidth]{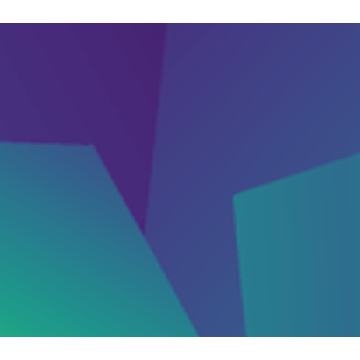}
    \includegraphics[width=\textwidth]{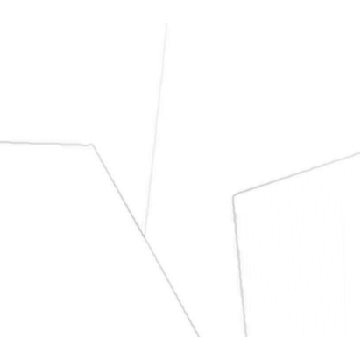}
    {\tiny (f) Ferstl~\etal~\cite{ferstl15}}
  \end{minipage}
  \begin{minipage}{0.24\textwidth} \centering
    \includegraphics[width=\textwidth]{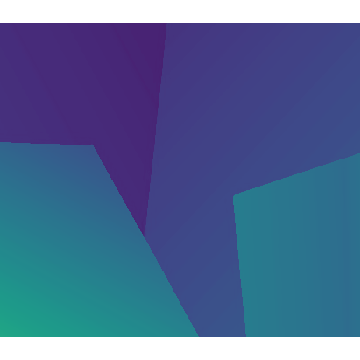}
    \includegraphics[width=\textwidth]{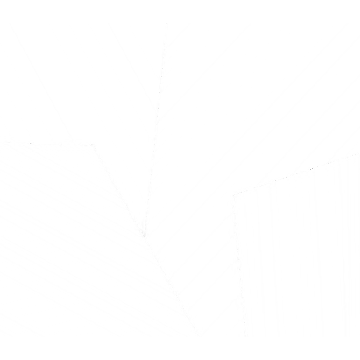}
    {\tiny (g) CNN only}
  \end{minipage}
  \begin{minipage}{0.24\textwidth} \centering
    \includegraphics[width=\textwidth]{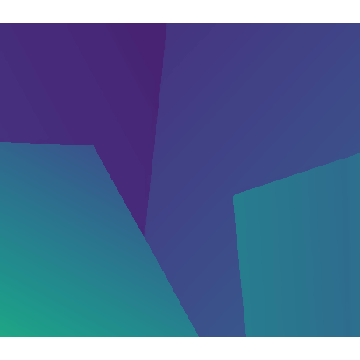}
    \includegraphics[width=\textwidth]{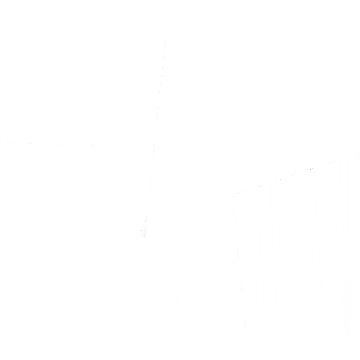}
    {\tiny (h) ATGV-Net}
  \end{minipage}
  
  \caption{
    Qualitative results for the noise-free Middlebury dataset sample \emph{Venus} and an upsampling factor $\scale = \times 4$.
    In (a) we show the ground-truth high-resolution depth map along with the low-resolution input in the top left corner, preserving the relative resolution.
    In (b) to (f) we visualize the results of bicubic upsampling and \sota approaches with the corresponding error images.
    In (g) we show the result of our network only, and in (h) we depict the result of our proposed \emph{ATGV-Net}.
    Best viewed magnified in the electronic version.
  }
  \label{fig:qual_results_mb2x4_03}
\end{figure}

\begin{figure}[b]
  \centering
  \begin{minipage}{0.24\textwidth} \centering
    \includegraphics[width=\textwidth]{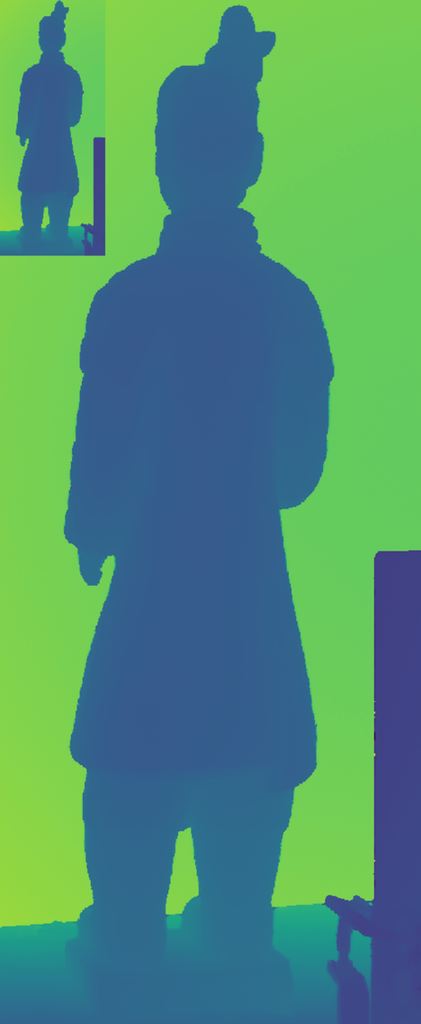}
    \mbox{\phantom{\includegraphics[width=\textwidth]{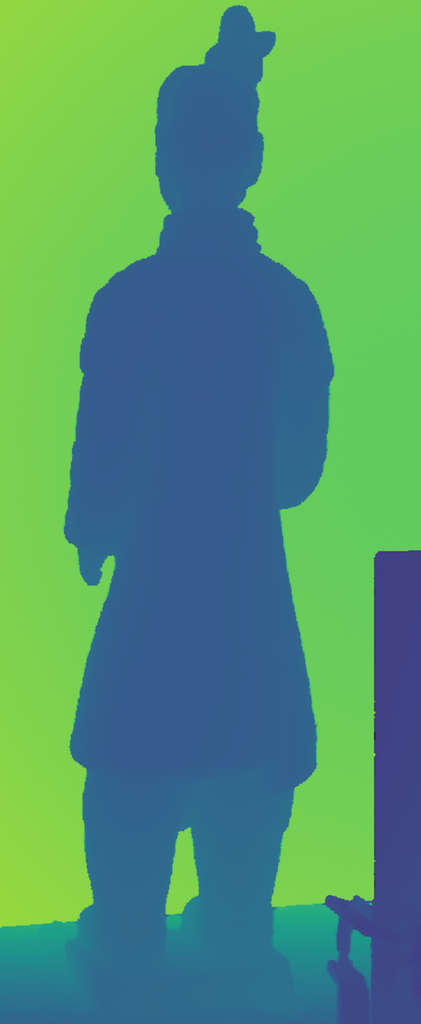}}}
    {\tiny (a) Input \& GT}
  \end{minipage}
  \begin{minipage}{0.24\textwidth} \centering
    \includegraphics[width=\textwidth]{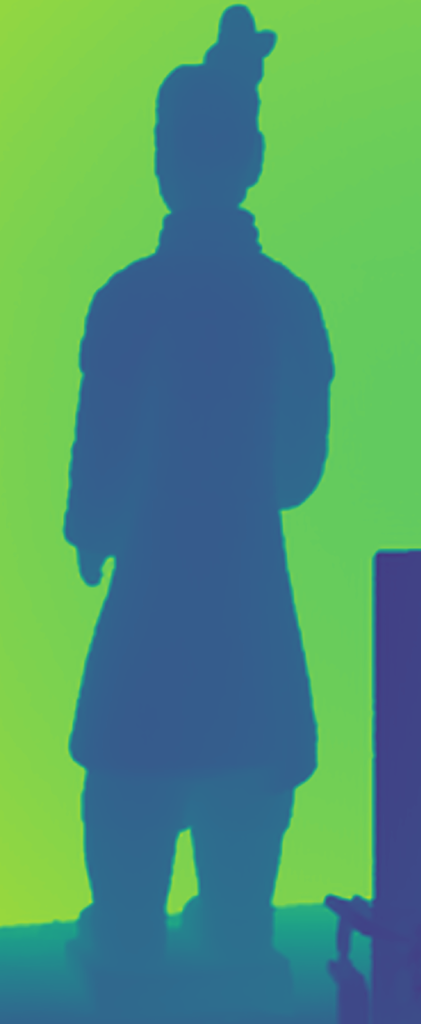}
    \includegraphics[width=\textwidth]{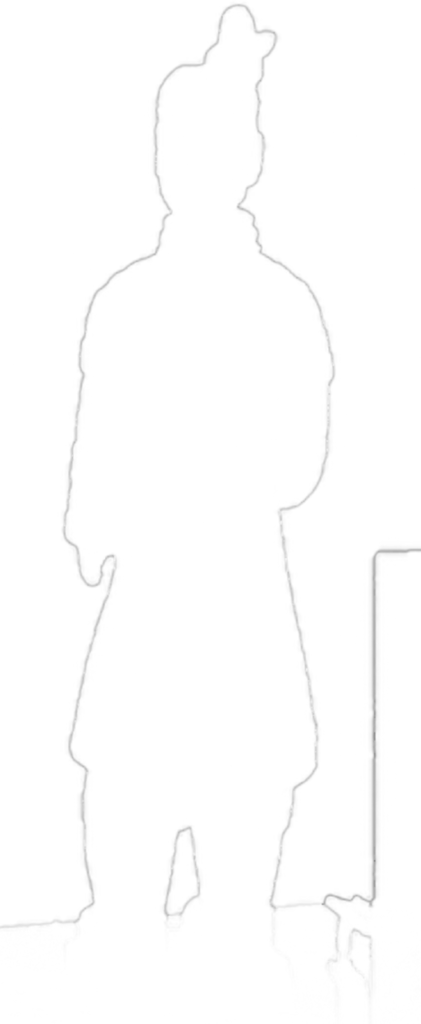}
    {\tiny (b) Ferstl~\etal~\cite{ferstl15}}
  \end{minipage}
  \begin{minipage}{0.24\textwidth} \centering
    \includegraphics[width=\textwidth]{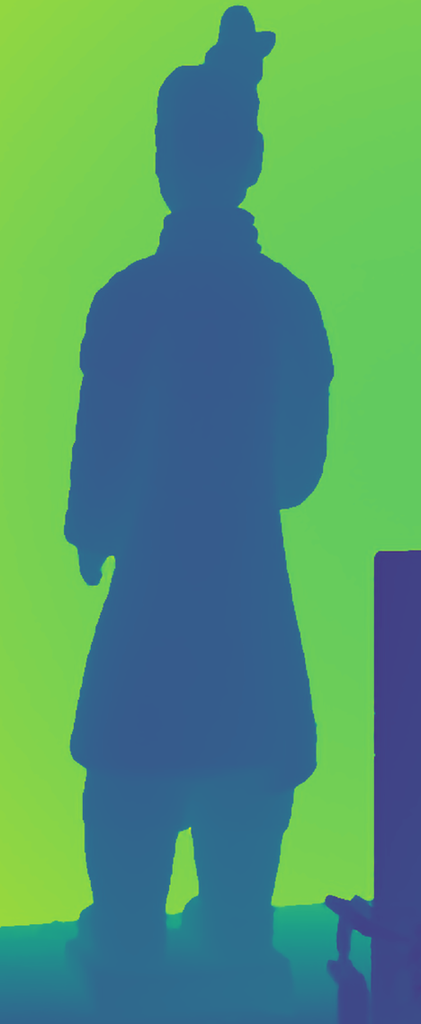}
    \includegraphics[width=\textwidth]{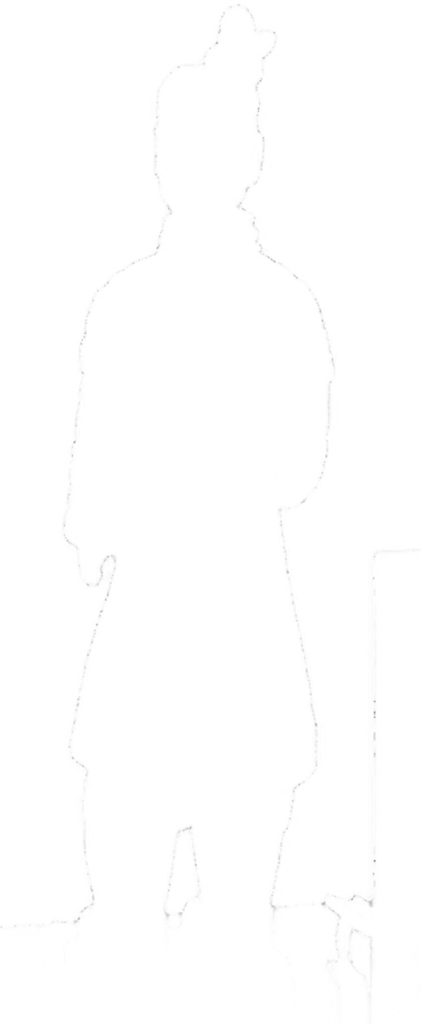}
    {\tiny (c) CNN only}
  \end{minipage}
  \begin{minipage}{0.24\textwidth} \centering
    \includegraphics[width=\textwidth]{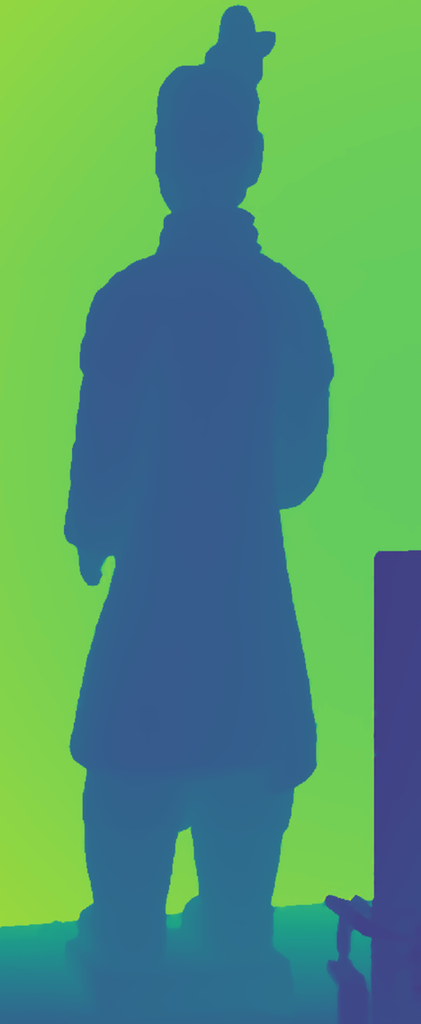}
    \includegraphics[width=\textwidth]{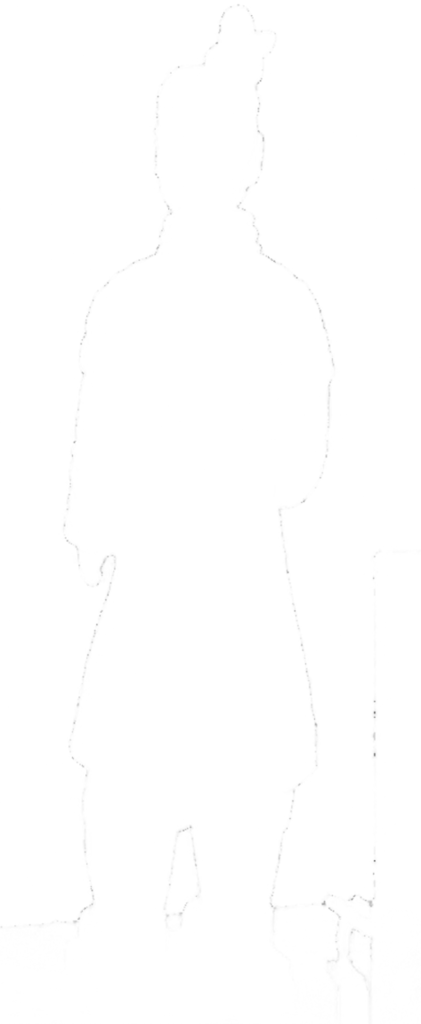}
    {\tiny (d) ATGV-Net}
  \end{minipage}
  
  \caption{
    Qualitative results for the Laserscan dataset sample \emph{Scan42}.
    In (a) we show the ground-truth high-resolution depth map along with the low-resolution input in the top left corner, preserving the relative resolution.
    In (b) we with visualize the results of a \sota approach and the corresponding error image.
    In (c) we show the result of our network only, and in (d) we depict the result of our proposed \emph{ATGV-Net}.
    Best viewed magnified in the electronic version.
  }
  \label{fig:qual_results_laserscan_02}
\end{figure}

\begin{figure}[b]
  \centering
  \begin{minipage}{0.24\textwidth} \centering
    \includegraphics[width=\textwidth]{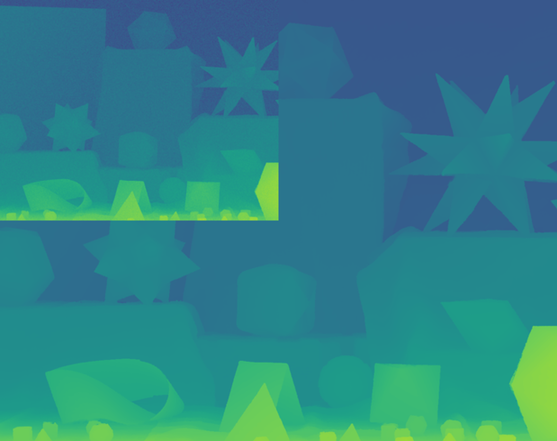}
    \mbox{\phantom{\includegraphics[width=\textwidth]{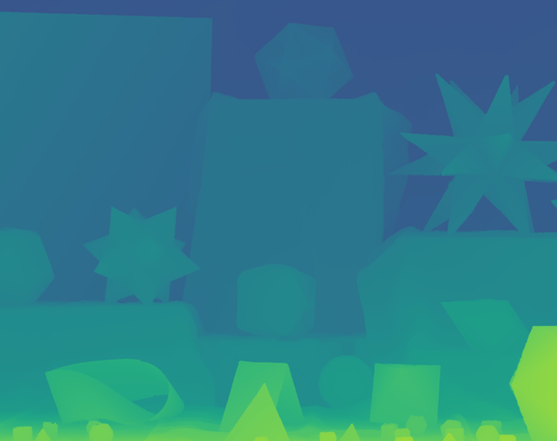}}}
    {\tiny (a) Input \& GT}
  \end{minipage}
  \begin{minipage}{0.24\textwidth} \centering
    \includegraphics[width=\textwidth]{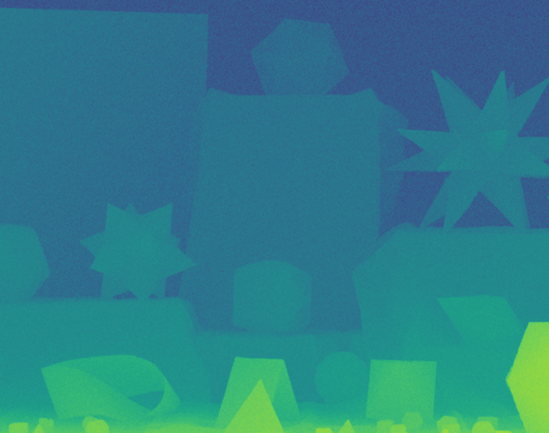}
    \includegraphics[width=\textwidth]{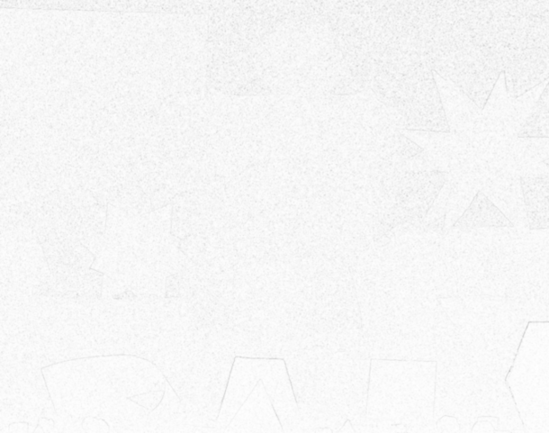}
    {\tiny (b) Bilinear}
  \end{minipage}
  \begin{minipage}{0.24\textwidth} \centering
    \includegraphics[width=\textwidth]{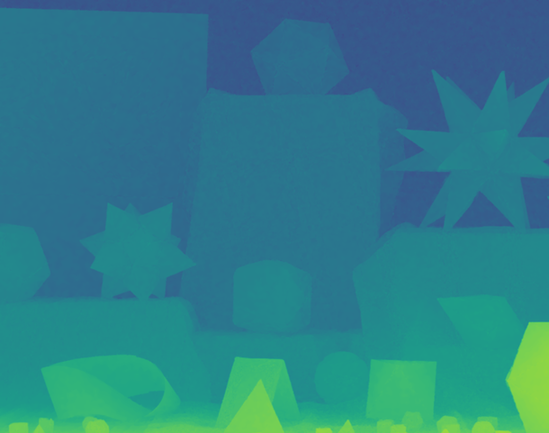}
    \includegraphics[width=\textwidth]{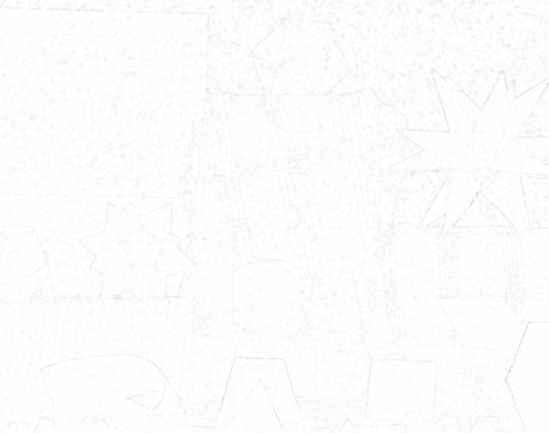}
    {\tiny (c) He~\etal~\cite{he10}}
  \end{minipage}
  \begin{minipage}{0.24\textwidth} \centering
    \includegraphics[width=\textwidth]{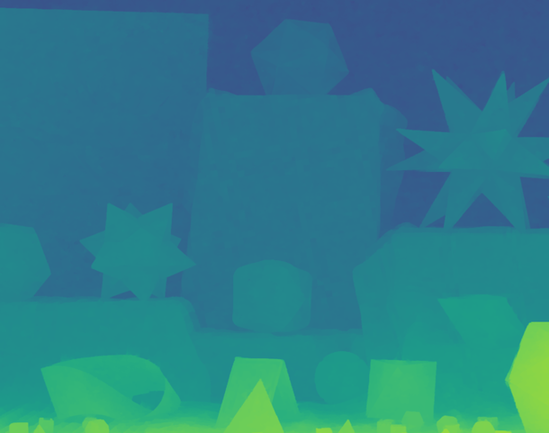}
    \includegraphics[width=\textwidth]{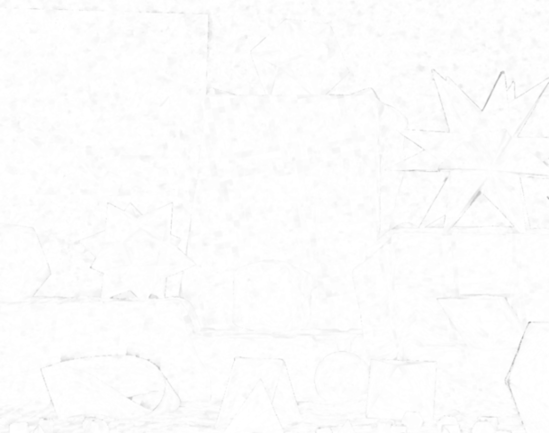}
    {\tiny (d) Park~\etal~\cite{park11}}
  \end{minipage}
  \begin{minipage}{0.24\textwidth} \centering
    \includegraphics[width=\textwidth]{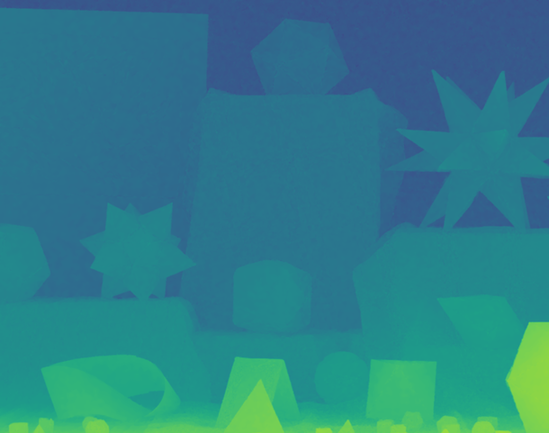}
    \includegraphics[width=\textwidth]{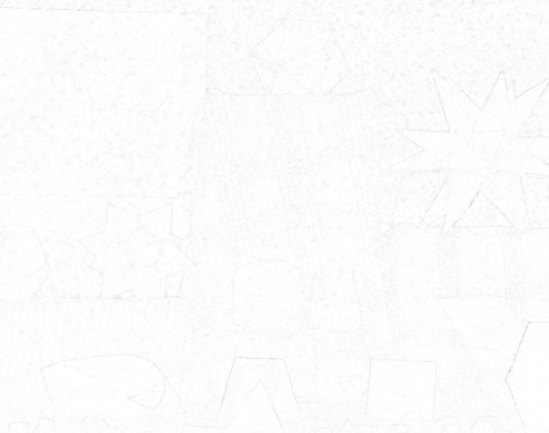}
    {\tiny (e) Yang~\etal~\cite{yang07}}
  \end{minipage}
  \begin{minipage}{0.24\textwidth} \centering
    \includegraphics[width=\textwidth]{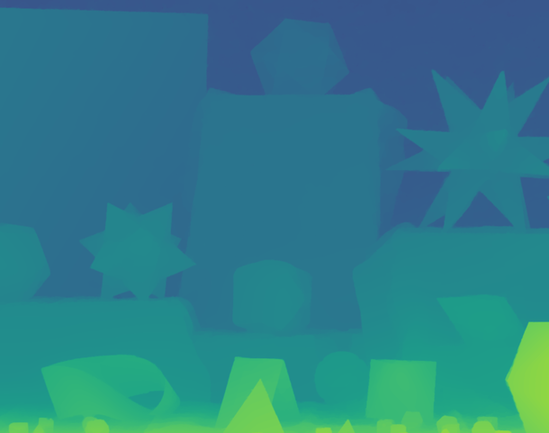}
    \includegraphics[width=\textwidth]{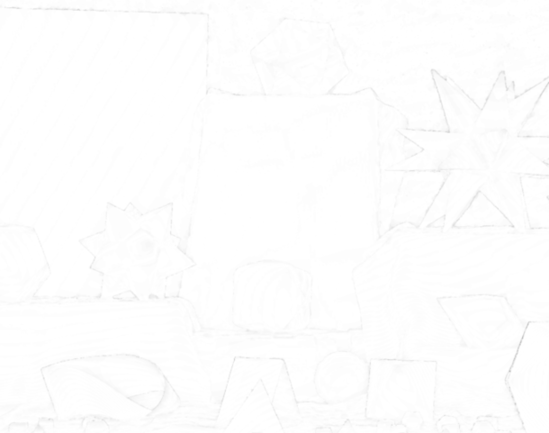}
    {\tiny (f) Ferstl~\etal~\cite{ferstl13}}
  \end{minipage}
  \begin{minipage}{0.24\textwidth} \centering
    \includegraphics[width=\textwidth]{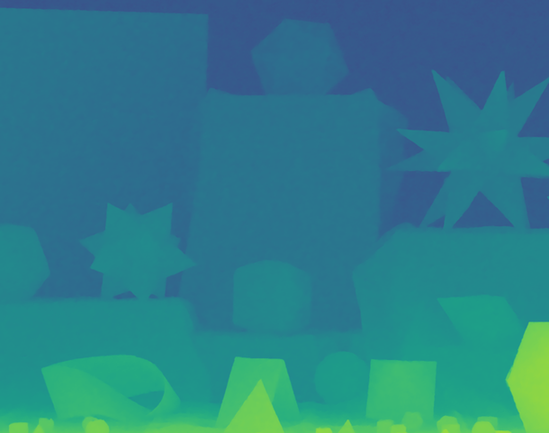}
    \includegraphics[width=\textwidth]{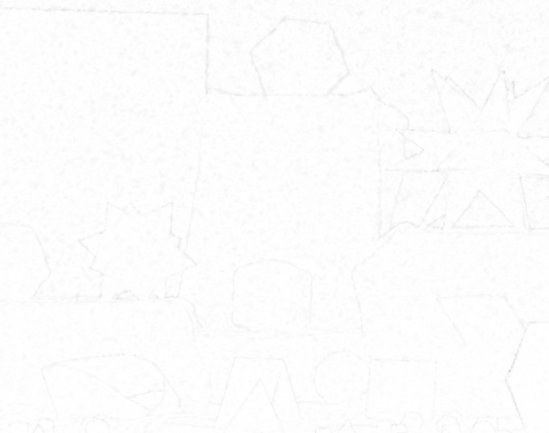}
    {\tiny (g) CNN only}
  \end{minipage}
  \begin{minipage}{0.24\textwidth} \centering
    \includegraphics[width=\textwidth]{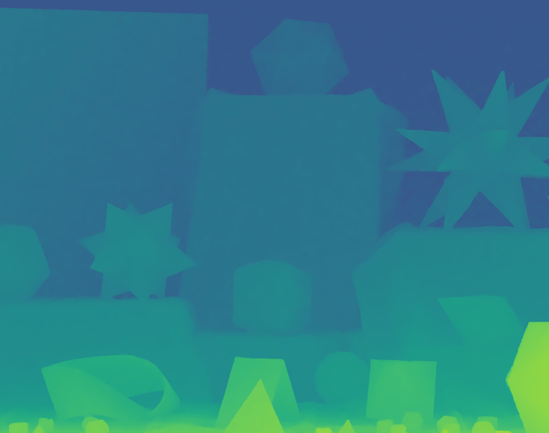}
    \includegraphics[width=\textwidth]{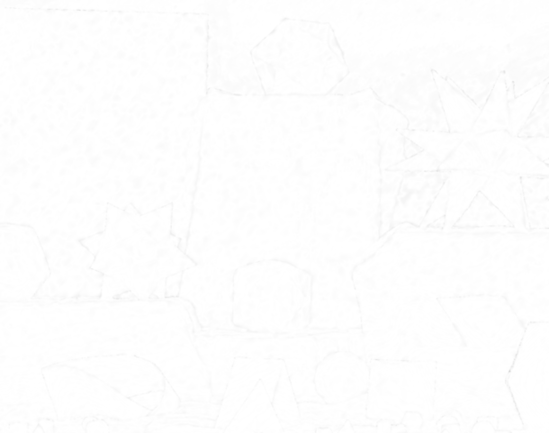}
    {\tiny (h) ATGV-Net}
  \end{minipage}
  
  \caption{
    Qualitative results for the noisy Middlebury dataset sample \emph{Moebius} and an upsampling factor $\scale = \times 2$.
    In (a) we show the ground-truth high-resolution depth map along with the low-resolution input in the top left corner, preserving the relative resolution.
    In (b) to (f) we visualize the results of bilinear upsampling and \sota approaches with the corresponding error images.
    In (g) we show the result of our network only, and in (h) we depict the result of our proposed \emph{ATGV-Net}.
    Best viewed magnified in the electronic version.
  }
  \label{fig:qual_results_nmbx2_02}
\end{figure}

\begin{figure}[b]
  \centering
  \begin{minipage}{0.24\textwidth} \centering
    \includegraphics[width=\textwidth]{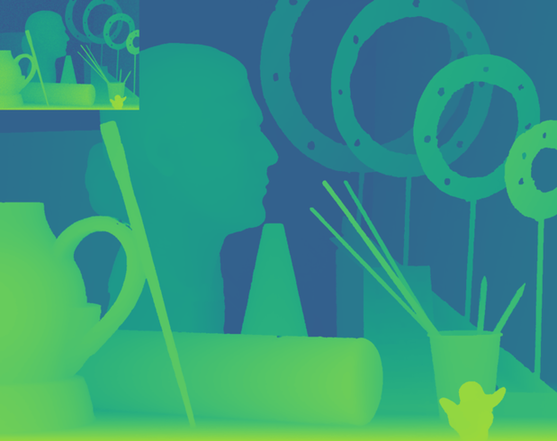}
    \mbox{\phantom{\includegraphics[width=\textwidth]{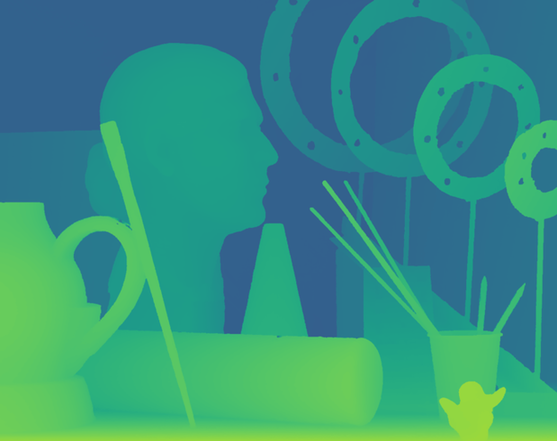}}}
    {\tiny (a) Input \& GT}
  \end{minipage}
  \begin{minipage}{0.24\textwidth} \centering
    \includegraphics[width=\textwidth]{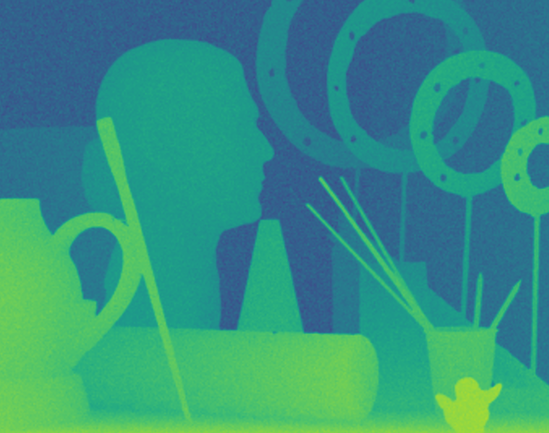}
    \includegraphics[width=\textwidth]{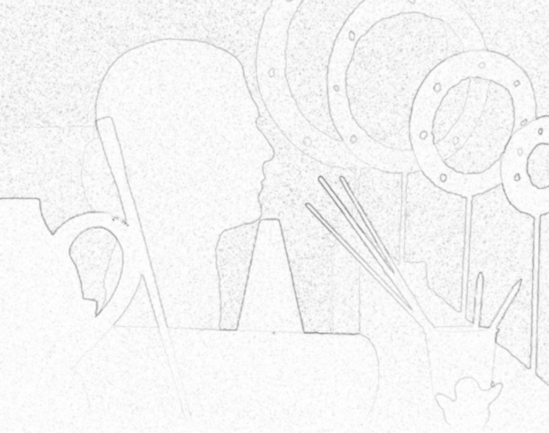}
    {\tiny (b) Bilinear}
  \end{minipage}
  \begin{minipage}{0.24\textwidth} \centering
    \includegraphics[width=\textwidth]{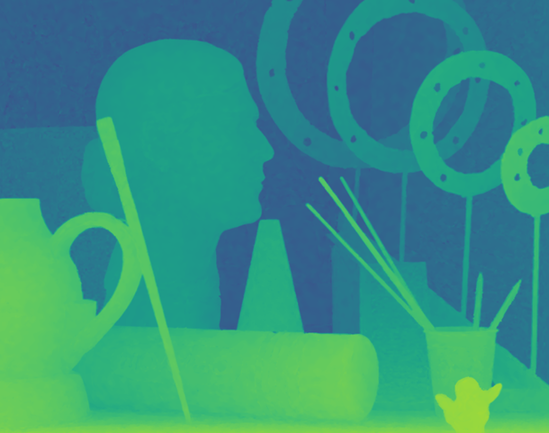}
    \includegraphics[width=\textwidth]{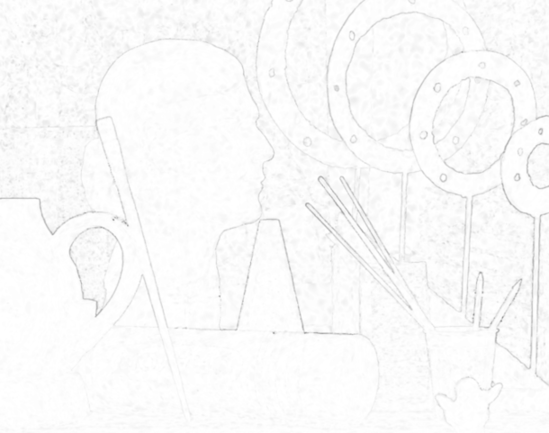}
    {\tiny (c) He~\etal~\cite{he10}}
  \end{minipage}
  \begin{minipage}{0.24\textwidth} \centering
    \includegraphics[width=\textwidth]{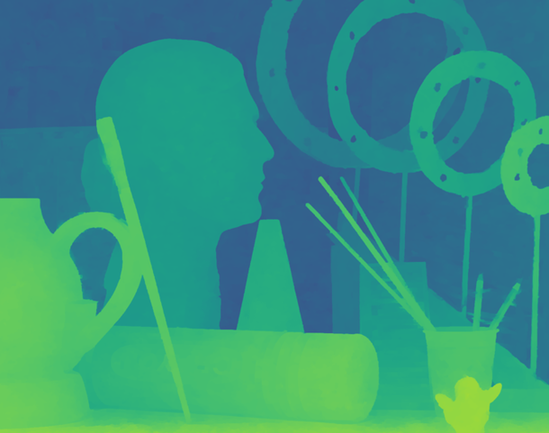}
    \includegraphics[width=\textwidth]{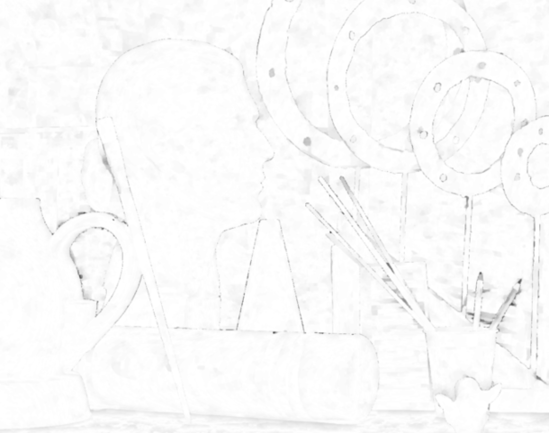}
    {\tiny (d) Park~\etal~\cite{park11}}
  \end{minipage}
  \begin{minipage}{0.24\textwidth} \centering
    \includegraphics[width=\textwidth]{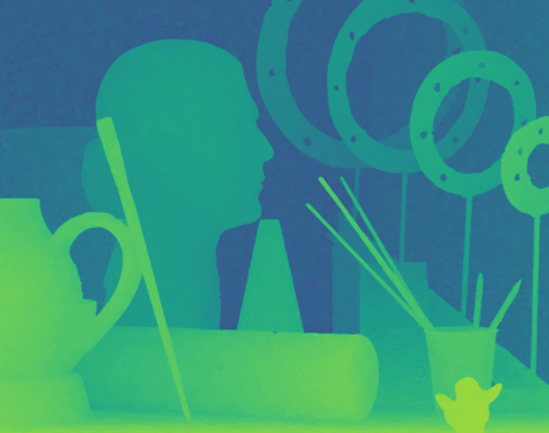}
    \includegraphics[width=\textwidth]{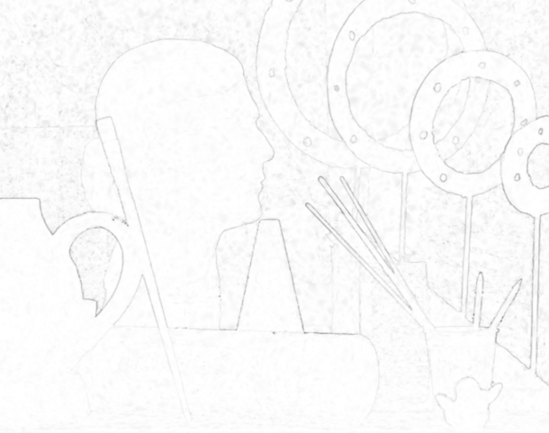}
    {\tiny (e) Yang~\etal~\cite{yang07}}
  \end{minipage}
  \begin{minipage}{0.24\textwidth} \centering
    \includegraphics[width=\textwidth]{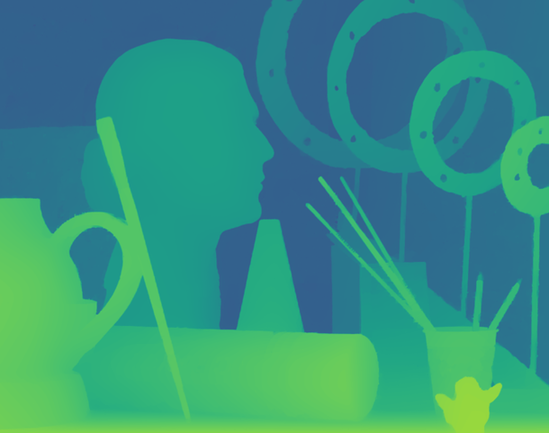}
    \includegraphics[width=\textwidth]{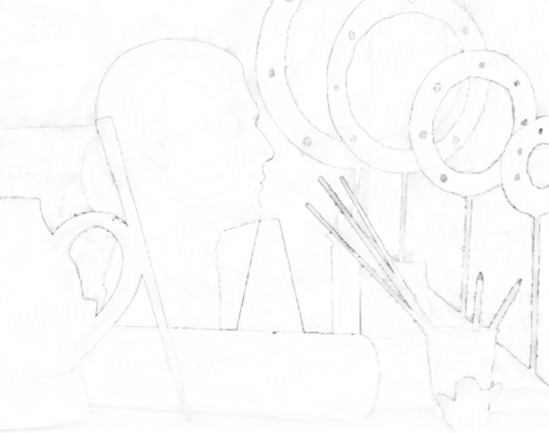}
    {\tiny (f) Ferstl~\etal~\cite{ferstl13}}
  \end{minipage}
  \begin{minipage}{0.24\textwidth} \centering
    \includegraphics[width=\textwidth]{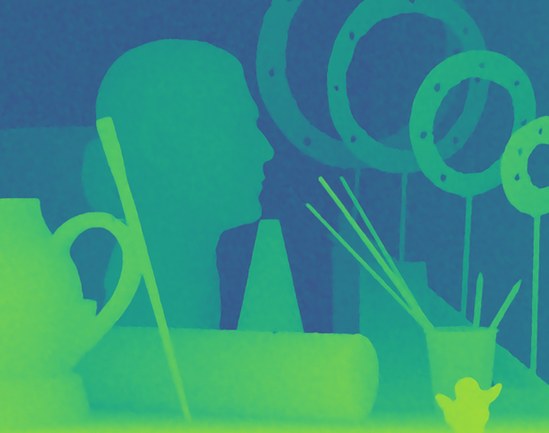}
    \includegraphics[width=\textwidth]{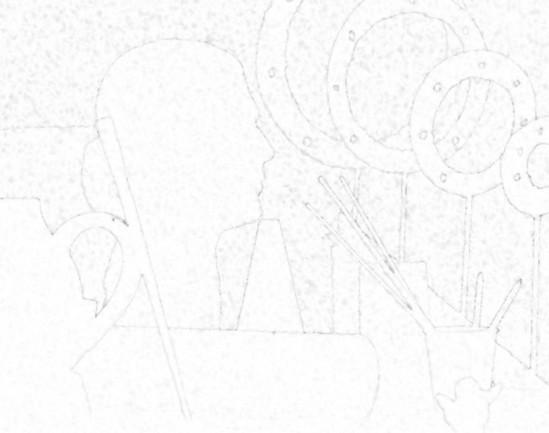}
    {\tiny (g) CNN only}
  \end{minipage}
  \begin{minipage}{0.24\textwidth} \centering
    \includegraphics[width=\textwidth]{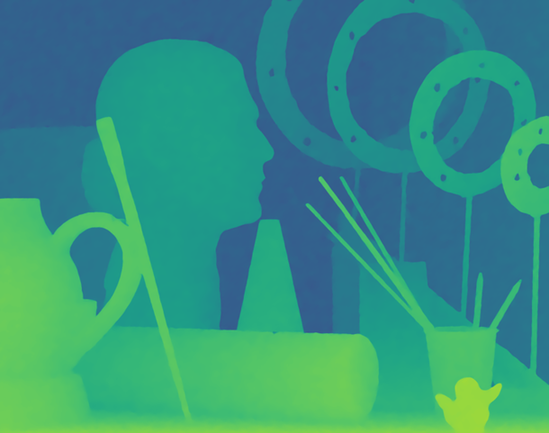}
    \includegraphics[width=\textwidth]{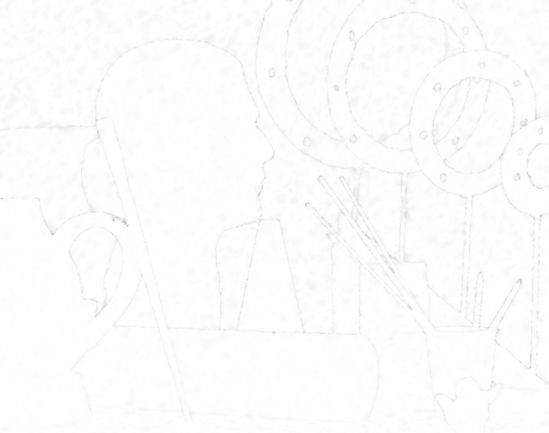}
    {\tiny (h) ATGV-Net}
  \end{minipage}
  
  \caption{
    Qualitative results for the noisy Middlebury dataset sample \emph{Art} and an upsampling factor $\scale = \times 4$.
    In (a) we show the ground-truth high-resolution depth map along with the low-resolution input in the top left corner, preserving the relative resolution.
    In (b) to (f) we visualize the results of bilinear upsampling and \sota approaches with the corresponding error images.
    In (g) we show the result of our network only, and in (h) we depict the result of our proposed \emph{ATGV-Net}.
    Best viewed magnified in the electronic version.
  }
  \label{fig:qual_results_nmbx4_00}
\end{figure}

\begin{figure}[b]
  \centering
  \begin{minipage}{0.24\textwidth} \centering
    \includegraphics[width=\textwidth]{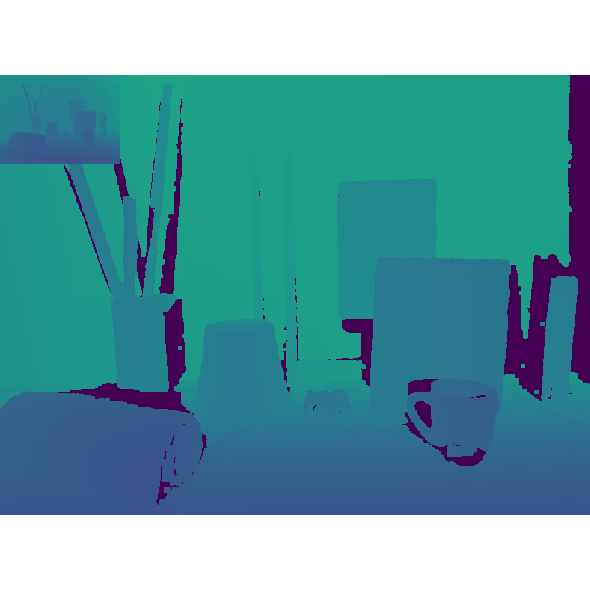}
    \mbox{\phantom{\includegraphics[width=\textwidth]{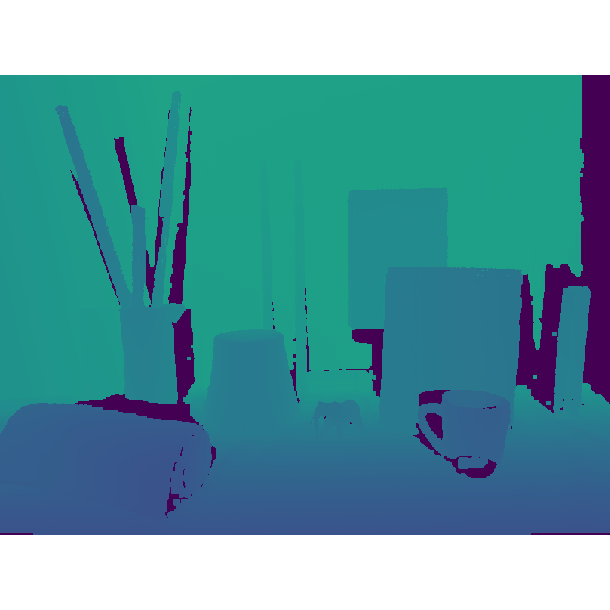}}}
    {\tiny (a) Input \& GT}
  \end{minipage}
  \begin{minipage}{0.24\textwidth} \centering
    \includegraphics[width=\textwidth]{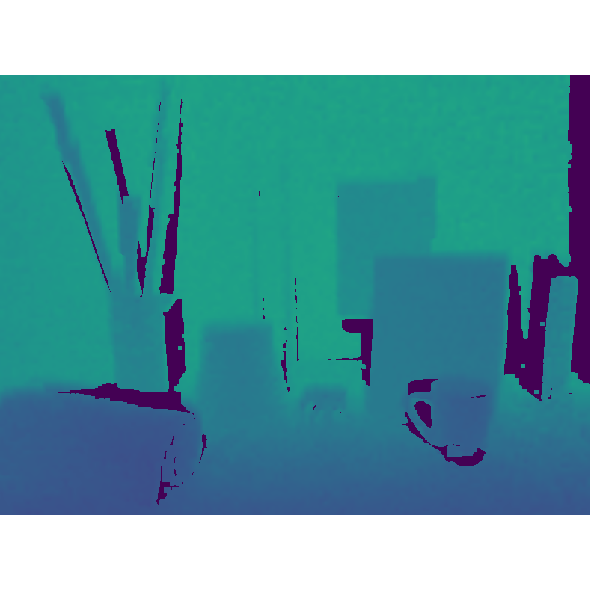}
    \includegraphics[width=\textwidth]{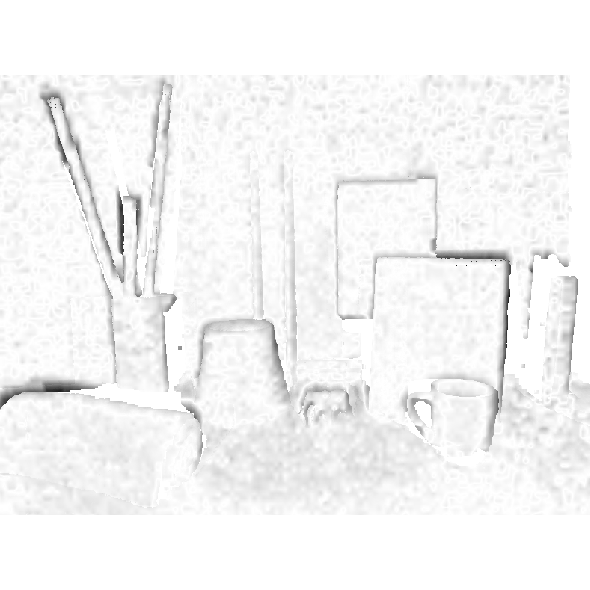}
    {\tiny (b) Bilinear}
  \end{minipage}
  \begin{minipage}{0.24\textwidth} \centering
    \includegraphics[width=\textwidth]{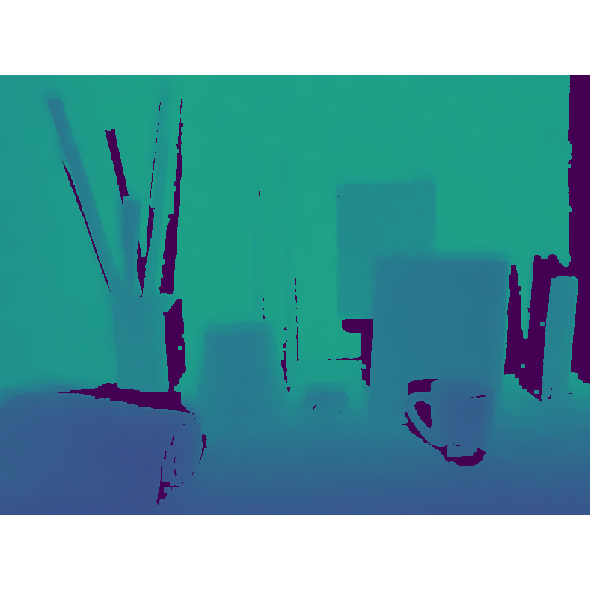}
    \includegraphics[width=\textwidth]{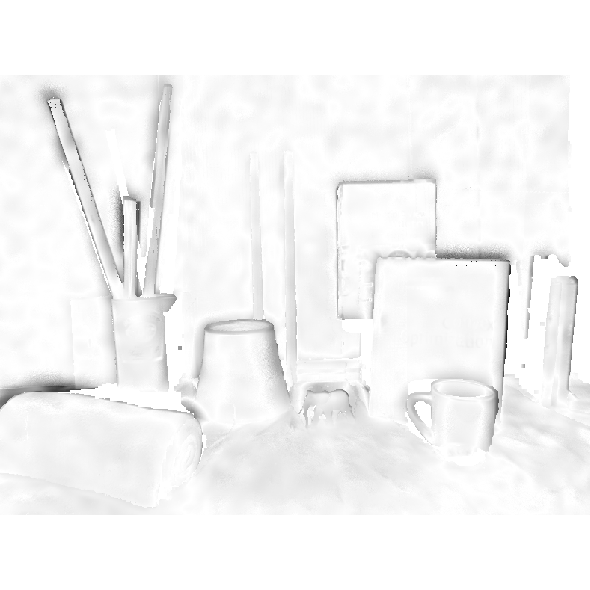}
    {\tiny (c) He~\etal~\cite{he10}}
  \end{minipage}
  \begin{minipage}{0.24\textwidth} \centering
    \includegraphics[width=\textwidth]{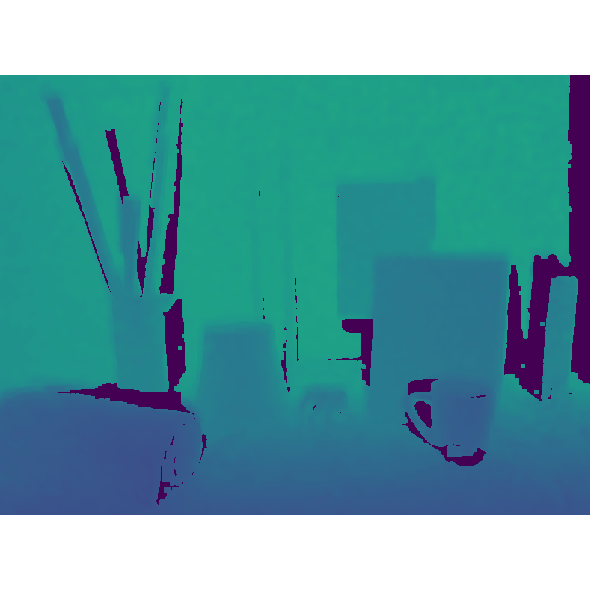}
    \includegraphics[width=\textwidth]{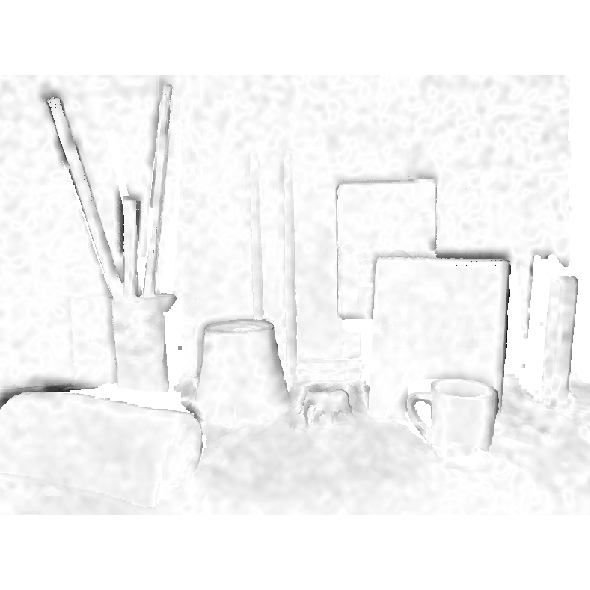}
    {\tiny (d) Kopf~\etal~\cite{kopf07}}
  \end{minipage}
  \begin{minipage}{0.24\textwidth} \centering
    \includegraphics[width=\textwidth]{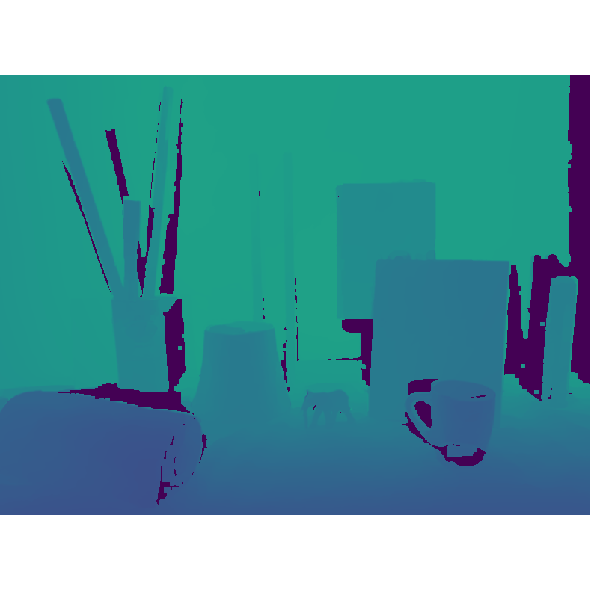}
    \includegraphics[width=\textwidth]{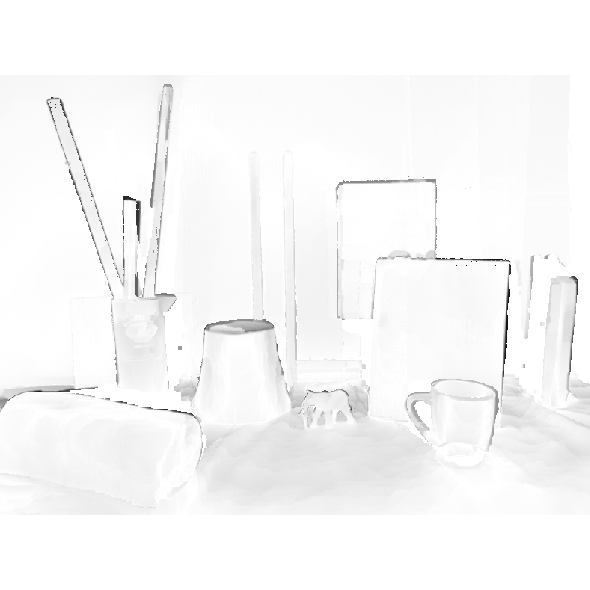}
    {\tiny (e) Ferstl~\etal~\cite{ferstl13}}
  \end{minipage}
  \begin{minipage}{0.24\textwidth} \centering
    \includegraphics[width=\textwidth]{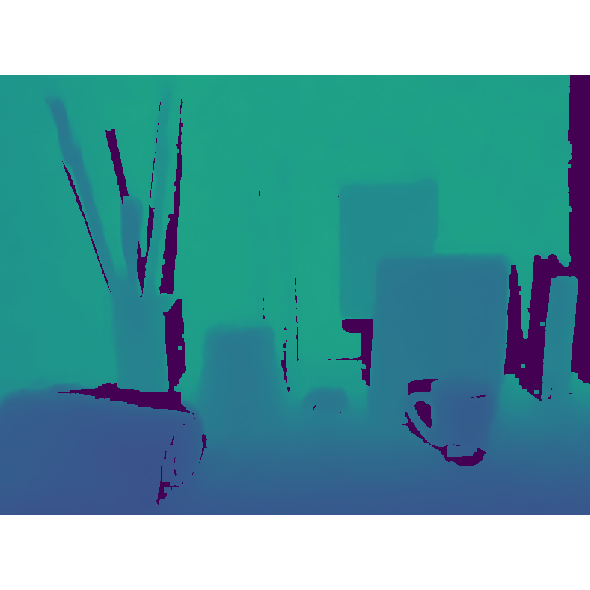}
    \includegraphics[width=\textwidth]{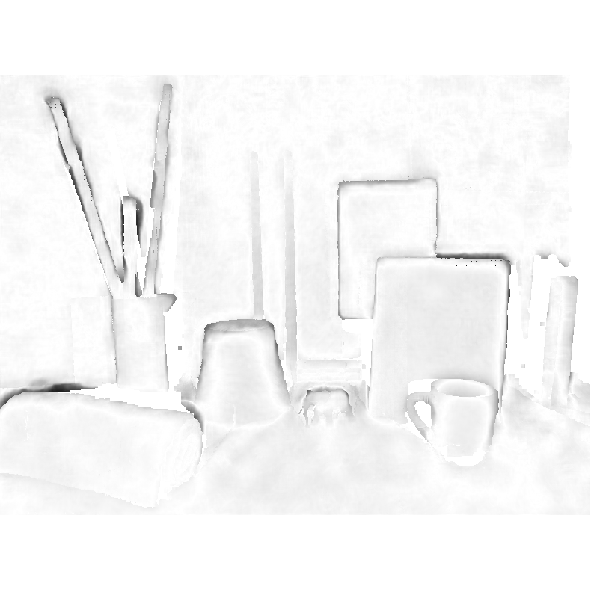}
    {\tiny (f) CNN only}
  \end{minipage}
  \begin{minipage}{0.24\textwidth} \centering
    \includegraphics[width=\textwidth]{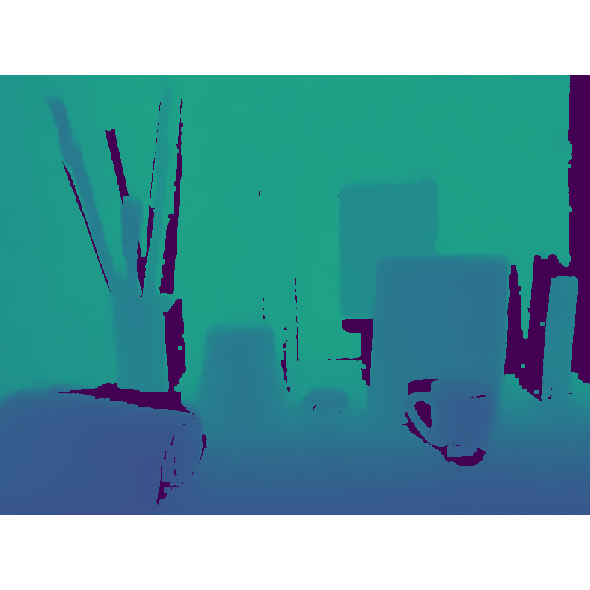}
    \includegraphics[width=\textwidth]{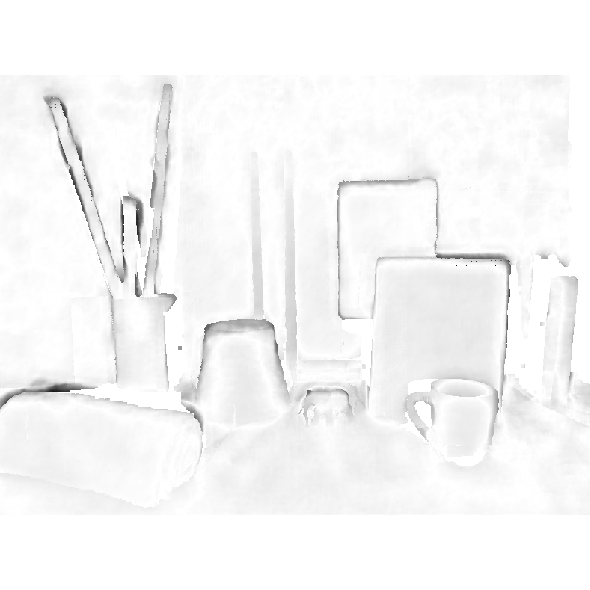}
    {\tiny (g) ATGV-Net}
  \end{minipage}
  \begin{minipage}{0.24\textwidth} \centering
    \mbox{\phantom{
      \includegraphics[width=\textwidth]{results/tofmark/atgvl2_srcnn10_bn_depth_00.png}
      \includegraphics[width=\textwidth]{results/tofmark/atgvl2_srcnn10_bn_depth_err_00.png}
      {\tiny (h) Phantom}
    }}
  \end{minipage}
  
  \caption{
    Qualitative results for the ToFMark dataset sample \emph{Books}.
    In (a) we show the ground-truth high-resolution depth map along with the low-resolution input in the top left corner, preserving the relative resolution.
    In (b) to (e) we visualize the results of bilinear upsampling and \sota approaches with the corresponding error images.
    In (f) we show the result of our network only, and in (g) we depict the result of our proposed \emph{ATGV-Net}.
    Best viewed magnified in the electronic version.
  }
  \label{fig:qual_results_tofmark_00}
\end{figure}

\begin{figure}[b]
  \centering
  \begin{minipage}{0.24\textwidth} \centering
    \includegraphics[width=\textwidth]{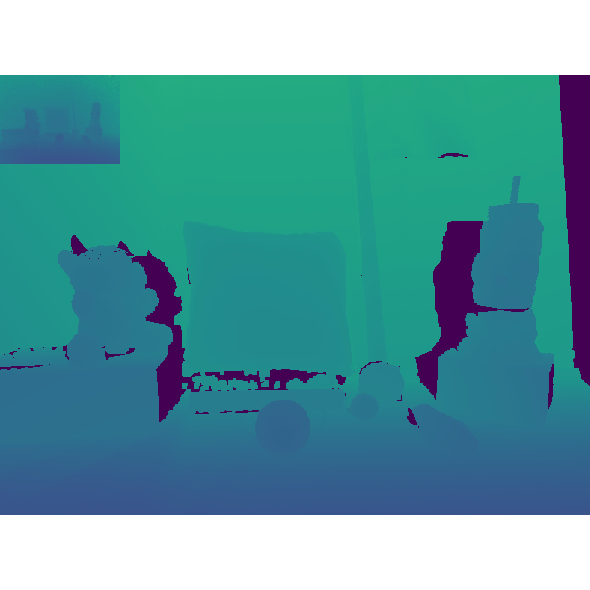}
    \mbox{\phantom{\includegraphics[width=\textwidth]{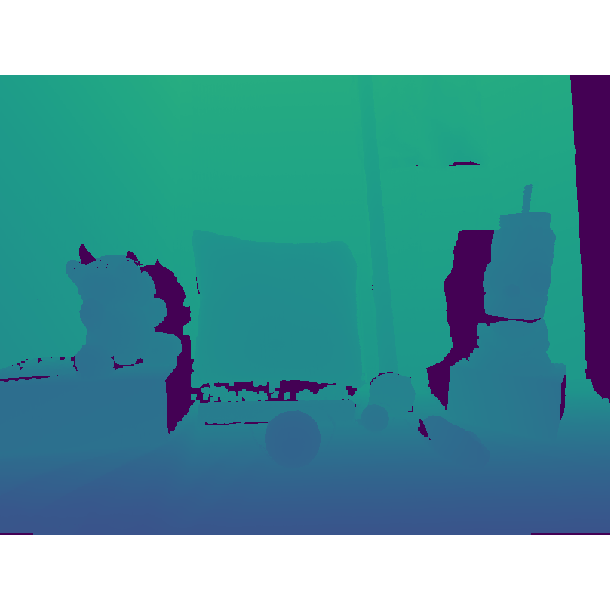}}}
    {\tiny (a) Input \& GT}
  \end{minipage}
  \begin{minipage}{0.24\textwidth} \centering
    \includegraphics[width=\textwidth]{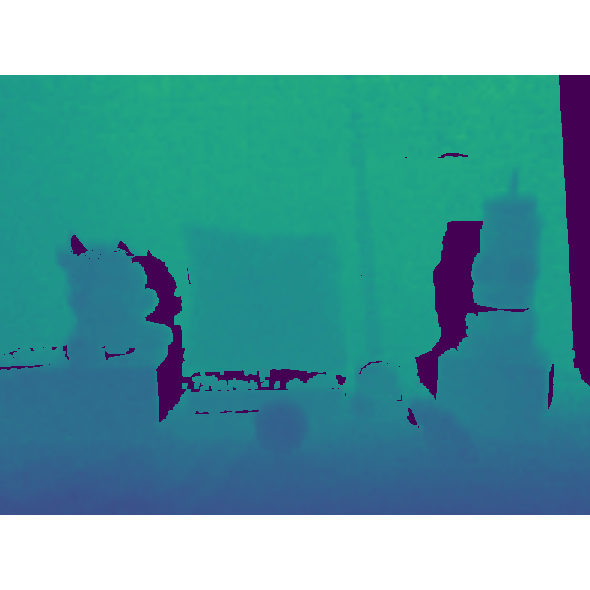}
    \includegraphics[width=\textwidth]{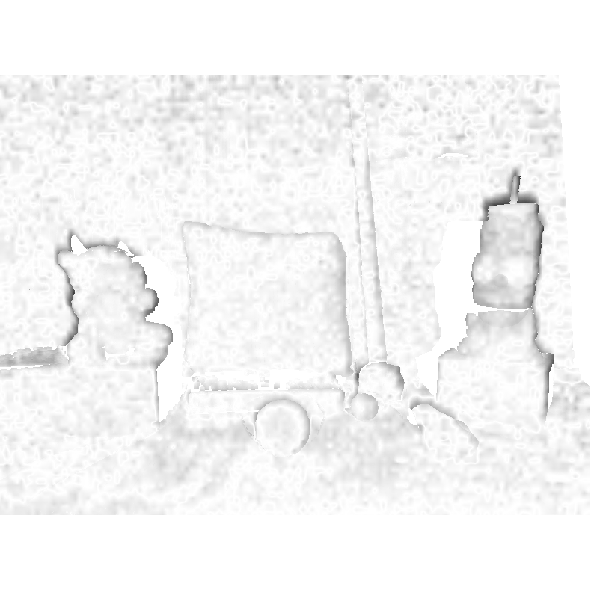}
    {\tiny (b) Bilinear}
  \end{minipage}
  \begin{minipage}{0.24\textwidth} \centering
    \includegraphics[width=\textwidth]{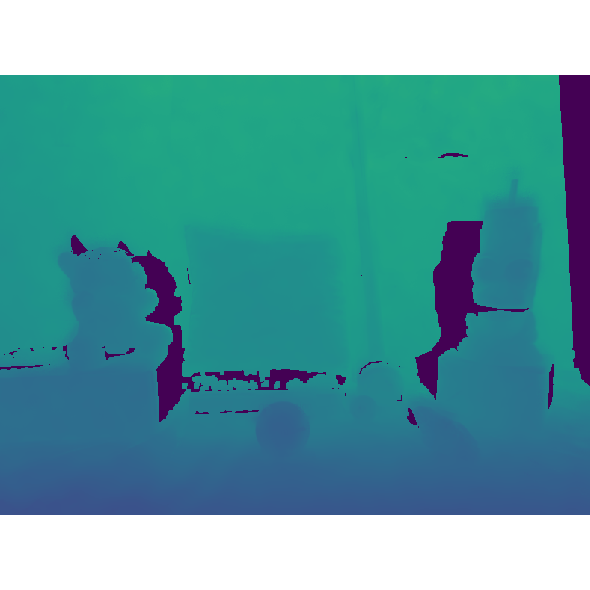}
    \includegraphics[width=\textwidth]{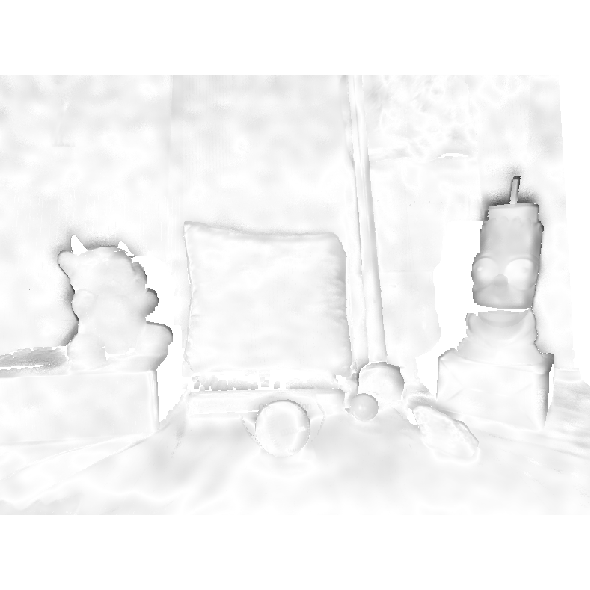}
    {\tiny (c) He~\etal~\cite{he10}}
  \end{minipage}
  \begin{minipage}{0.24\textwidth} \centering
    \includegraphics[width=\textwidth]{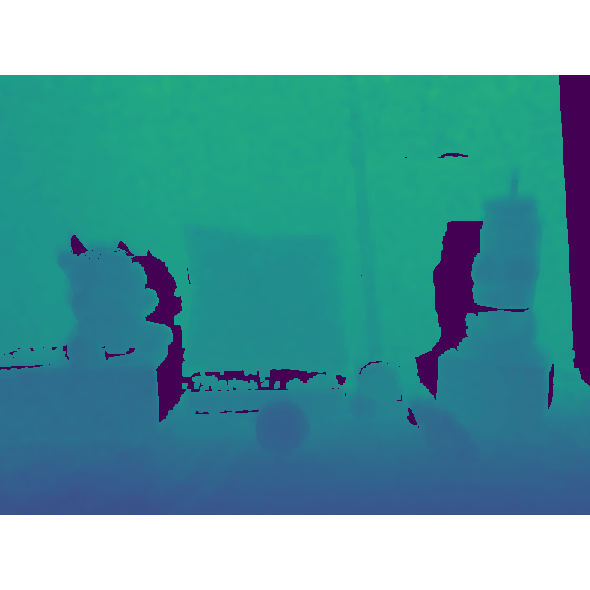}
    \includegraphics[width=\textwidth]{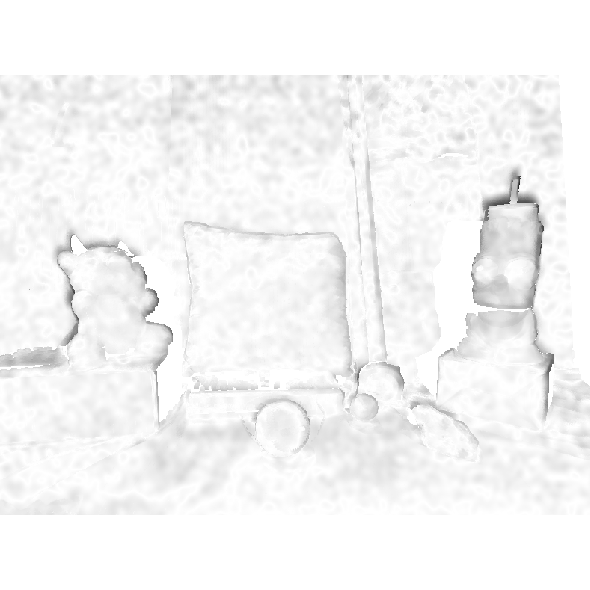}
    {\tiny (d) Kopf~\etal~\cite{kopf07}}
  \end{minipage}
  \begin{minipage}{0.24\textwidth} \centering
    \includegraphics[width=\textwidth]{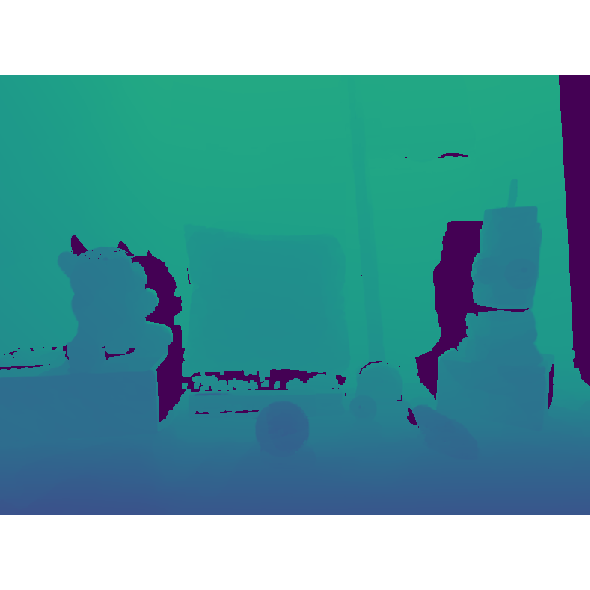}
    \includegraphics[width=\textwidth]{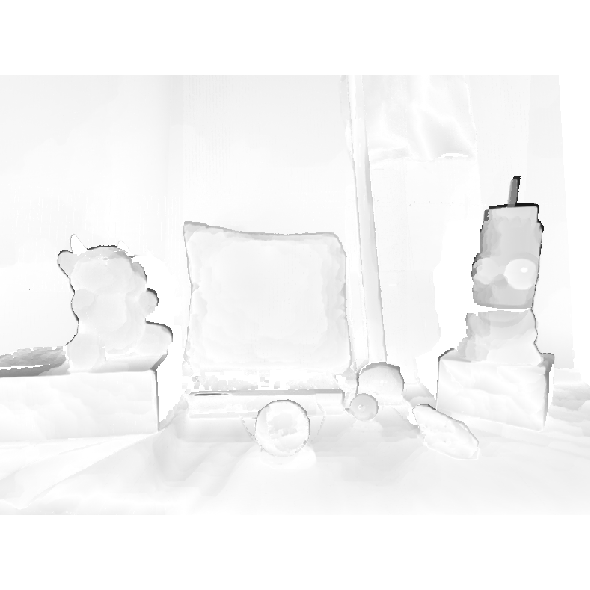}
    {\tiny (e) Ferstl~\etal~\cite{ferstl13}}
  \end{minipage}
  \begin{minipage}{0.24\textwidth} \centering
    \includegraphics[width=\textwidth]{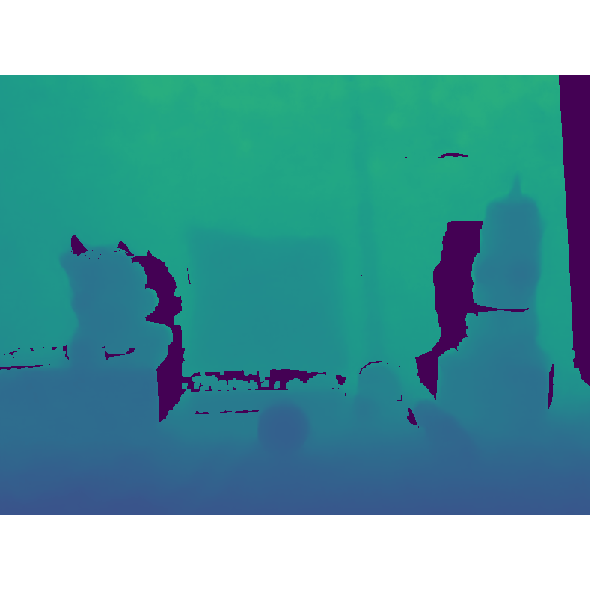}
    \includegraphics[width=\textwidth]{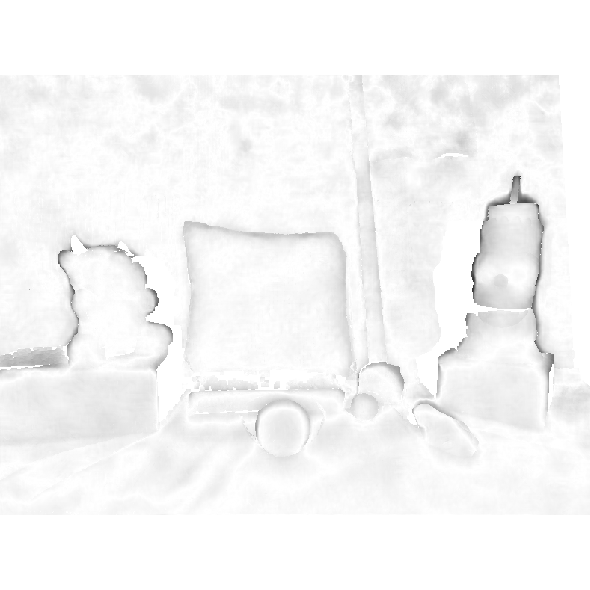}
    {\tiny (f) CNN only}
  \end{minipage}
  \begin{minipage}{0.24\textwidth} \centering
    \includegraphics[width=\textwidth]{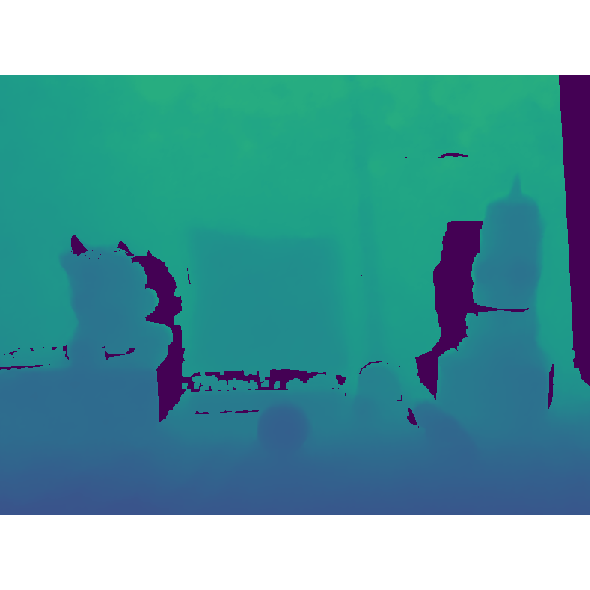}
    \includegraphics[width=\textwidth]{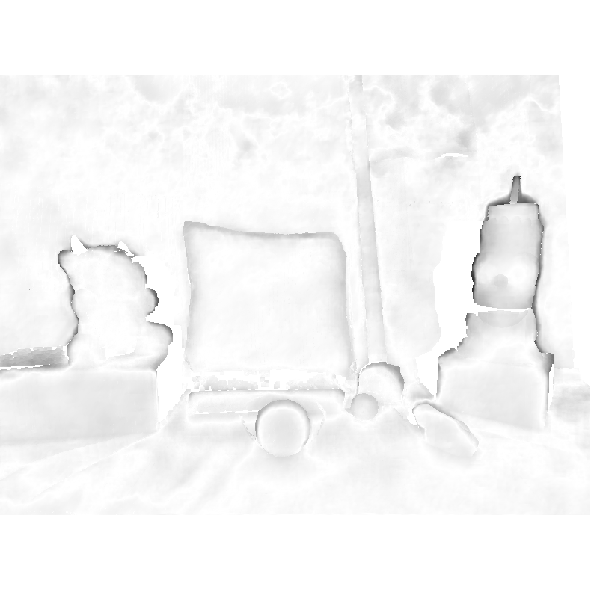}
    {\tiny (g) ATGV-Net}
  \end{minipage}
  \begin{minipage}{0.24\textwidth} \centering
    \mbox{\phantom{
      \includegraphics[width=\textwidth]{results/tofmark/atgvl2_srcnn10_bn_depth_01.png}
      \includegraphics[width=\textwidth]{results/tofmark/atgvl2_srcnn10_bn_depth_err_01.png}
      {\tiny (h) Phantom}
    }}
  \end{minipage}
  
  \caption{
    Qualitative results for the ToFMark dataset sample \emph{Devil}.
    In (a) we show the ground-truth high-resolution depth map along with the low-resolution input in the top left corner, preserving the relative resolution.
    In (b) to (e) we visualize the results of bilinear upsampling and \sota approaches with the corresponding error images.
    In (f) we show the result of our network only, and in (g) we depict the result of our proposed \emph{ATGV-Net}.
    Best viewed magnified in the electronic version.
  }
  \label{fig:qual_results_tofmark_01}
\end{figure}

\begin{figure}[b]
  \centering
  \begin{minipage}{0.24\textwidth} \centering
    \includegraphics[width=\textwidth]{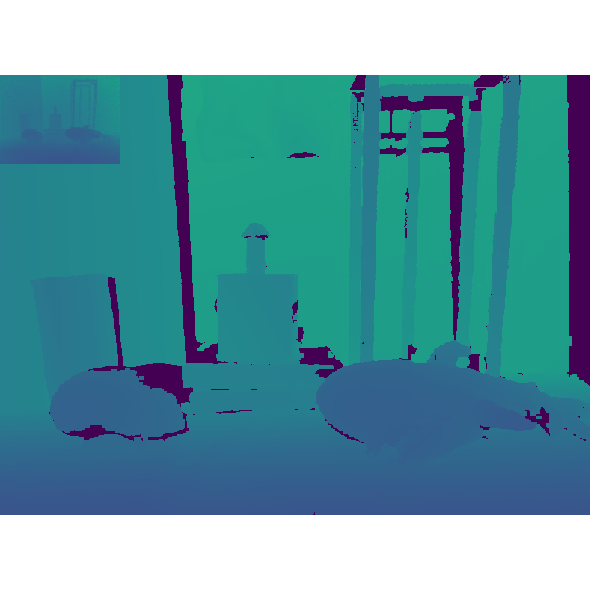}
    \mbox{\phantom{\includegraphics[width=\textwidth]{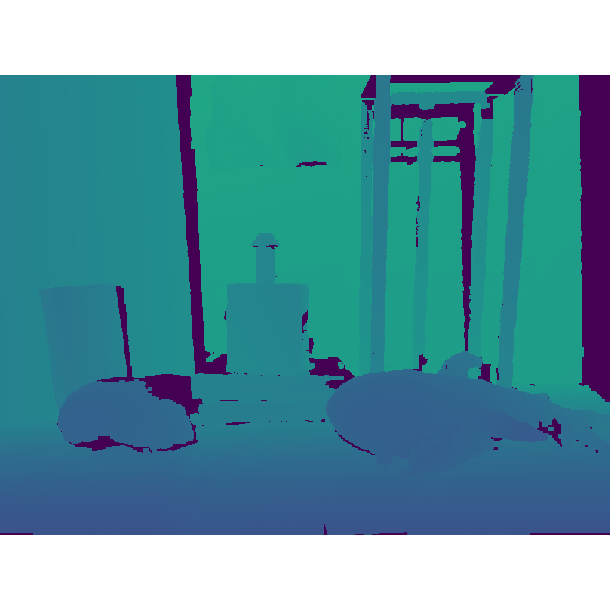}}}
    {\tiny (a) Input \& GT}
  \end{minipage}
  \begin{minipage}{0.24\textwidth} \centering
    \includegraphics[width=\textwidth]{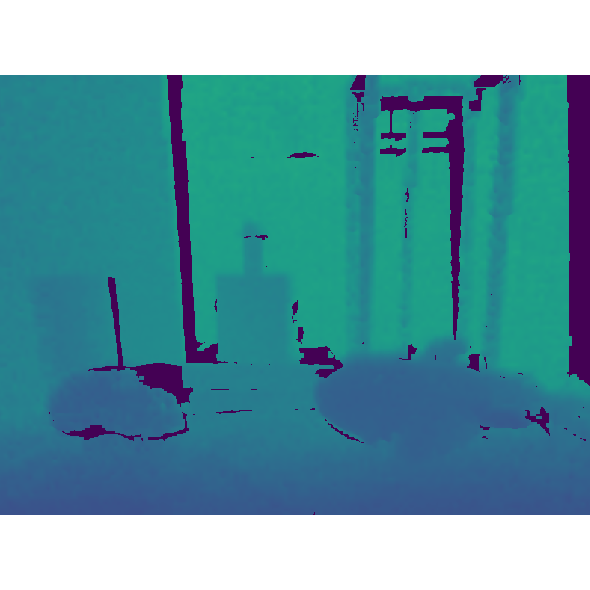}
    \includegraphics[width=\textwidth]{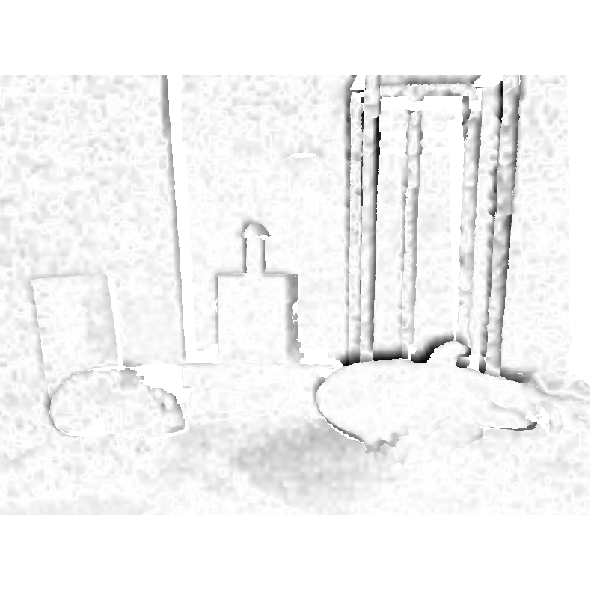}
    {\tiny (b) Bilinear}
  \end{minipage}
  \begin{minipage}{0.24\textwidth} \centering
    \includegraphics[width=\textwidth]{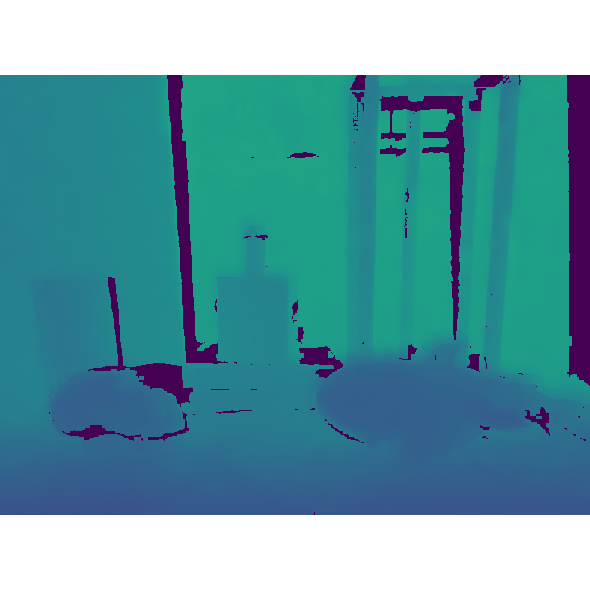}
    \includegraphics[width=\textwidth]{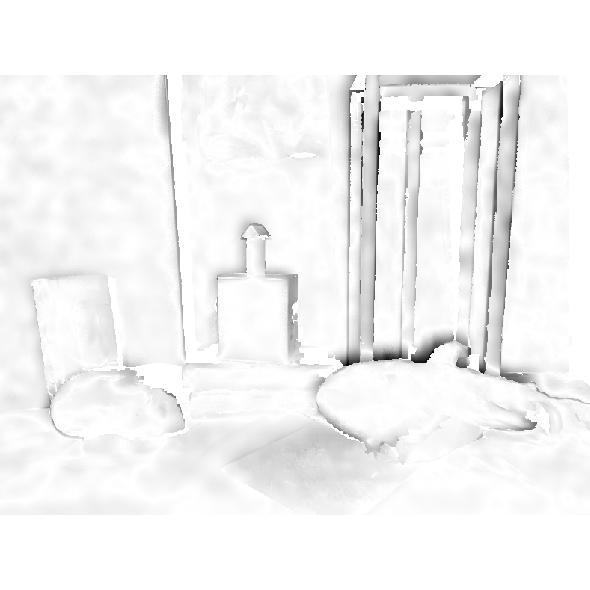}
    {\tiny (c) He~\etal~\cite{he10}}
  \end{minipage}
  \begin{minipage}{0.24\textwidth} \centering
    \includegraphics[width=\textwidth]{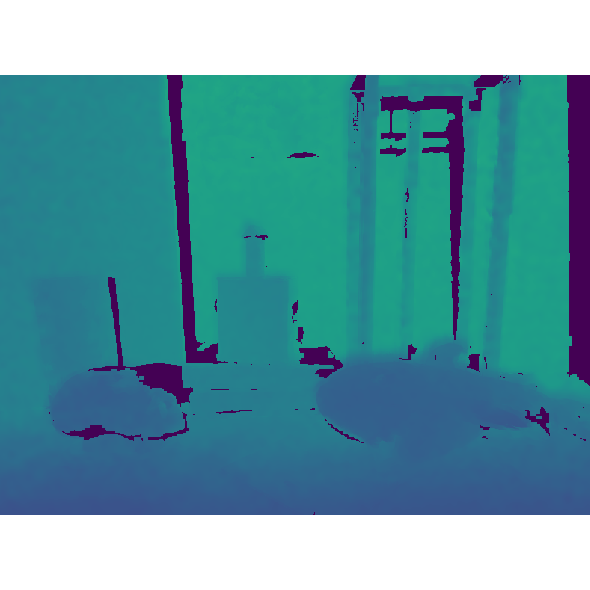}
    \includegraphics[width=\textwidth]{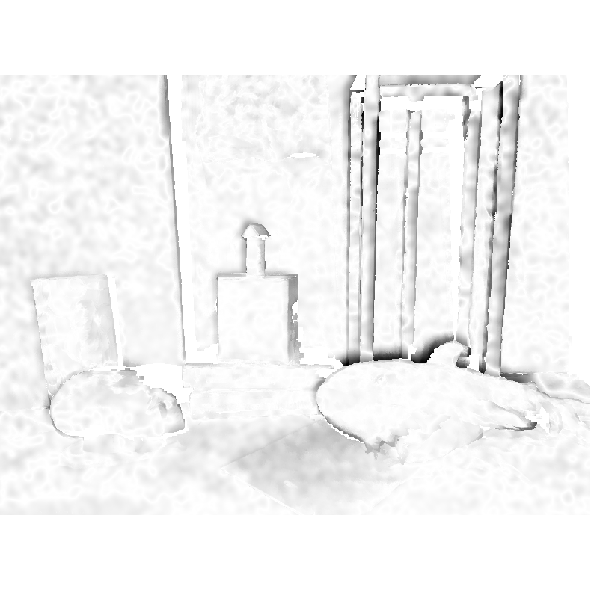}
    {\tiny (d) Kopf~\etal~\cite{kopf07}}
  \end{minipage}
  \begin{minipage}{0.24\textwidth} \centering
    \includegraphics[width=\textwidth]{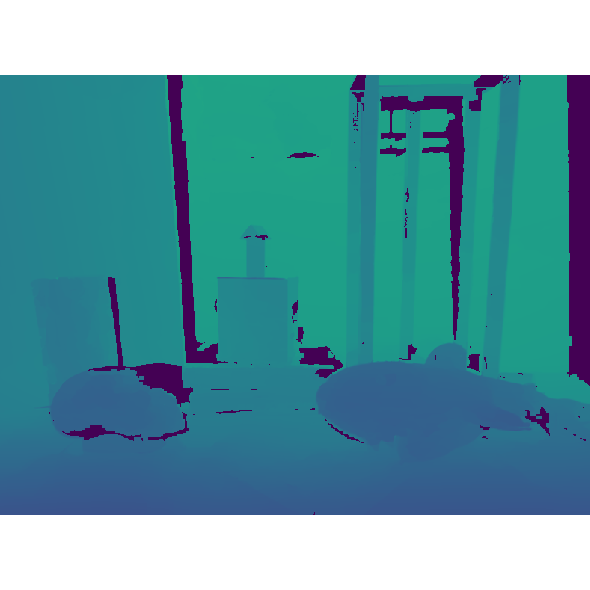}
    \includegraphics[width=\textwidth]{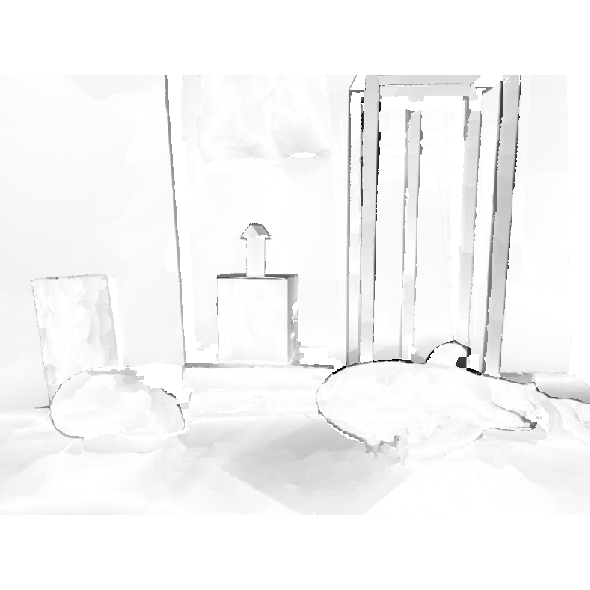}
    {\tiny (e) Ferstl~\etal~\cite{ferstl13}}
  \end{minipage}
  \begin{minipage}{0.24\textwidth} \centering
    \includegraphics[width=\textwidth]{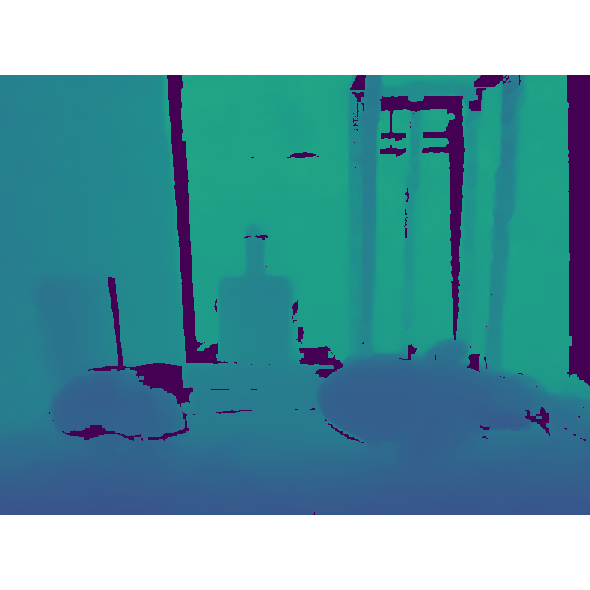}
    \includegraphics[width=\textwidth]{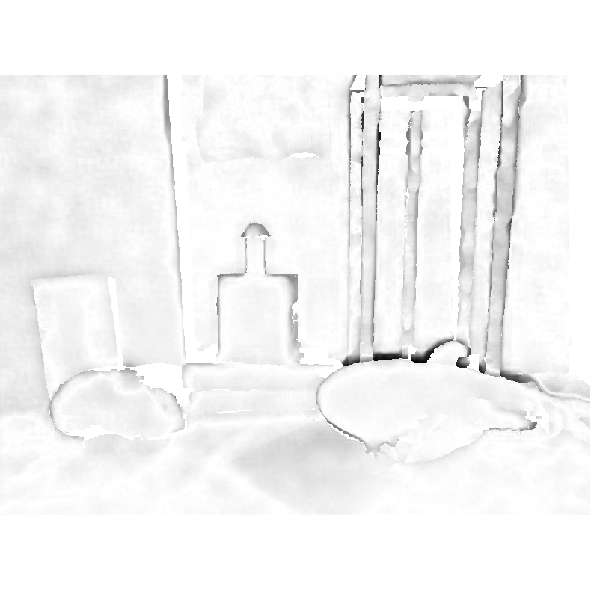}
    {\tiny (f) CNN only}
  \end{minipage}
  \begin{minipage}{0.24\textwidth} \centering
    \includegraphics[width=\textwidth]{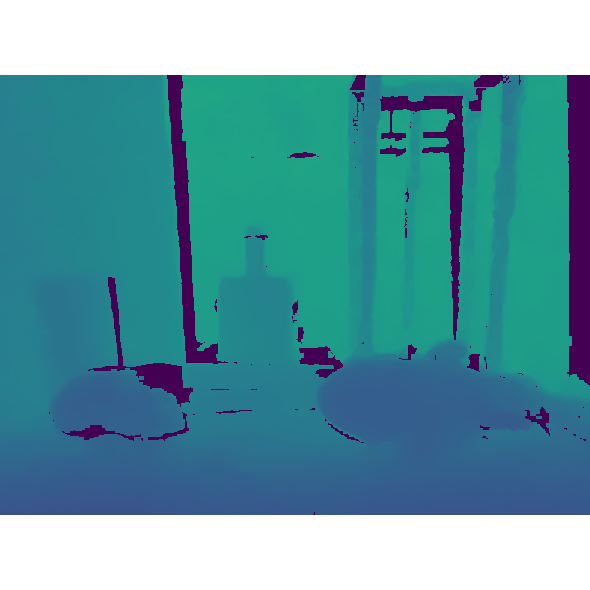}
    \includegraphics[width=\textwidth]{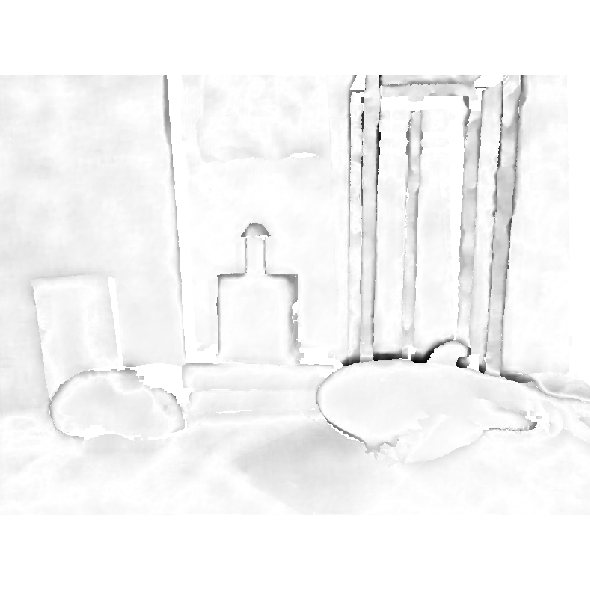}
    {\tiny (g) ATGV-Net}
  \end{minipage}
  \begin{minipage}{0.24\textwidth} \centering
    \mbox{\phantom{
      \includegraphics[width=\textwidth]{results/tofmark/atgvl2_srcnn10_bn_depth_02.png}
      \includegraphics[width=\textwidth]{results/tofmark/atgvl2_srcnn10_bn_depth_err_02.png}
      {\tiny (h) Phantom}
    }}
  \end{minipage}
  
  \caption{
    Qualitative results for the ToFMark dataset sample \emph{Shark}.
    In (a) we show the ground-truth high-resolution depth map along with the low-resolution input in the top left corner, preserving the relative resolution.
    In (b) to (e) we visualize the results of bilinear upsampling and \sota approaches with the corresponding error images.
    In (f) we show the result of our network only, and in (g) we depict the result of our proposed \emph{ATGV-Net}.
    Best viewed magnified in the electronic version.
  }
  \label{fig:qual_results_tofmark_02}
\end{figure}

In Section~\ref{sec:quantitative_results} we present additional quantitative results in terms of mean absolute error.
In Section~\ref{sec:qualitative_results} we depict the qualitative results of four different benchmark datasets and different upsampling factors $\scale$ with additional error images.
In Figures~\ref{fig:qual_results_mb2x2_00} and \ref{fig:qual_results_mb2x2_01} we present the qualitative results for the Middlebury images \emph{Cones} and \emph{Teddy} for an upsampling factor of $\times 2$.
Similarly, we show the results on the same dataset for the images \emph{Teddy}, \emph{Tsukuba} and \emph{Venus} and an upsamling factor of $\times 4$ in Figures~\ref{fig:qual_results_mb2x4_01} to \ref{fig:qual_results_mb2x4_03}.
A qualitative result of the Laserscan dataset is presented in Figures~\ref{fig:qual_results_laserscan_02}.
For the noisy Middlebury sample \emph{Moebius} and an upsampling factor of $\times 2$, we show the qualitative results in Figure~\ref{fig:qual_results_nmbx2_02}.
The results for \emph{Art} and an upsampling factor $\times 4$ on the same dataset are visualized in Figure~\ref{fig:qual_results_nmbx4_00}.
Finally, we depict the qualitative results for the challenging ToFMark dataset in Figures~\ref{fig:qual_results_tofmark_00} to \ref{fig:qual_results_tofmark_02}.

\newpage
\subsection{Computation Graph}\label{sec:computation_graph}
\clearpage

\newpage
\subsection{Additional Quantitative Results}\label{sec:quantitative_results}
In Table~\ref{tab:results_noisy_mb_sad} and Table~\ref{tab:results_tofmark_sad} we present our quantitative results in terms of mean absolute error (MAE) on the noisy Middlebury and ToFMark dataset, respectively.
Although we only compare to methods that utilize an additional guidance image and that our method was not trained on this metric, we obtain reasonable results.

\begin{table}[h]
  \centering
  \caption{
    Results on noisy Middlebury data.
    We report the error as MAE in pixel disparity and highlight the best result in boldface and the second best in italic.
  }
  \tiny{
    \begin{tabular}{l r r r r r r } \toprule
 & \multicolumn{3}{c}{$\times2$} & \multicolumn{3}{c}{$\times4$} \\ \cmidrule(lr){2-4}\cmidrule(lr){5-7}
 & Art & Books & Moebius & Art & Books & Moebius\\ \midrule
NN & $4.65$ & $4.30$ & $5.08$ & $5.01$ & $4.68$ & $5.20$\\ 
Bilinear & $3.09$ & $2.91$ & $3.21$ & $3.59$ & $3.12$ & $3.45$\\ 
Yang \etal \cite{yang07} & $1.36$ & $1.12$ & $1.25$ & $1.93$ & $1.47$ & $1.63$\\ 
He \etal \cite{he10} & $1.92$ & $1.60$ & $1.77$ & $2.40$ & $1.82$ & $2.03$\\ 
Diebel \& Thrun \cite{diebel05} & $1.62$ & $1.34$ & $1.47$ & $2.24$ & $2.08$ & $2.29$\\ 
Chan \etal \cite{chan08} & $1.83$ & $1.04$ & $1.17$ & $2.90$ & $1.36$ & $1.55$\\ 
Park \etal \cite{park11} & $1.24$ & $0.99$ & $1.03$ & $1.82$ & $1.43$ & $1.49$\\ 
Ferstl \etal \cite{ferstl13}  & $0.84$ & $\mathbf{0.51}$ & $\mathbf{0.57}$ & $\mathit{1.29}$ & $\mathbf{0.75}$ & $\mathbf{0.90}$\\  \midrule
CNN only & $0.95$ & $0.85$ & $1.00$ & $1.92$ & $1.65$ & $1.91$\\ 
CNN + ATGV-L2 & $\mathit{0.81}$ & $0.72$ & $0.86$ & $1.74$ & $1.49$ & $1.74$\\ 
ATGV-Net & $\mathbf{0.71}$ & $\mathit{0.69}$ & $\mathit{0.74}$ & $\mathbf{1.26}$ & $\mathit{1.04}$ & $\mathit{1.21}$\\ 
\bottomrule\end{tabular}

  }
  \label{tab:results_noisy_mb_sad}
\end{table}

\begin{table}[h]
  \centering
  \caption{
    Results on real Time-of-Flight data from the ToFMark benchmark dataset.
    We report the error as MAE in $mm$ and highlight the best result in boldface and the second best in italic.
  }
  \tiny{
    \begin{tabular}{l r r r } \toprule
 & Books & Devil & Shark\\ \midrule
NN & $18.21$ & $21.83$ & $19.36$\\ 
Bilinear & $17.10$ & $20.17$ & $18.66$\\ 
Kopf \etal \cite{kopf07} & $16.03$ & $18.79$ & $27.57$\\ 
He \etal \cite{he10} & $15.74$ & $18.21$ & $27.04$\\ 
Ferstl \etal \cite{ferstl13}  & $\mathbf{12.36}$ & $\mathit{15.29}$ & $\mathbf{14.68}$\\  \midrule
ATGV-Net & $\mathit{13.22}$ & $\mathbf{13.07}$ & $\mathit{15.61}$\\ 
\bottomrule\end{tabular}

  }
  \label{tab:results_tofmark_sad}
\end{table}

\newpage
\subsection{Qualitative Results}\label{sec:qualitative_results}

\clearpage

\clearpage

\clearpage

\clearpage

\clearpage

\clearpage

\clearpage

\clearpage

\clearpage

\clearpage

\end{document}